\documentclass{article}

\PassOptionsToPackage{numbers, compress}{natbib}

\usepackage[preprint]{nips_2018}

\usepackage[utf8]{inputenc} 
\usepackage[T1]{fontenc}   
\usepackage{hyperref}      
\usepackage{url}           
\usepackage{booktabs}      
\usepackage{amsfonts}      
\usepackage{nicefrac}     
\usepackage{microtype}    
\usepackage{graphicx}
\usepackage{subfigure}
\usepackage{multirow}
\usepackage{float}
\usepackage{caption}
\usepackage{colortbl,booktabs}
\usepackage{amsmath,amssymb}

\title{Large Margin Few-Shot Learning}

\author{Yong Wang$^{1}$, Xiao-Ming Wu$^{2}$\thanks{corresponding author} ,  Qimai Li$^{2}$, Jiatao Gu$^{1}$, \\
\textbf{Wangmeng Xiang$^{2}$, Lei Zhang$^{2}$, Victor O.K. Li$^{1}$}\\
$^{1}$Department of Electrical and Electronic Engineering, The University of Hong Kong\\
$^{2}$Department of Computing, The Hong Kong Polytechnic University \\
\texttt{\{wangyong, jiataogu, vli\}@eee.hku.hk,}\\
\texttt{\{xiao-ming.wu\}@polyu.edu.hk,}\\
\texttt{\{csqmli, cswxiang, cslzhang\}@comp.polyu.edu.hk} \\
}

\begin{document}

\maketitle

\begin{abstract}
	
The key issue of few-shot learning is learning to generalize. This paper proposes a large margin principle to improve the generalization capacity of metric based methods for few-shot learning. To realize it, we develop a unified framework to learn a more discriminative metric space by augmenting the classification loss function with a large margin distance loss function for training. Extensive experiments on two state-of-the-art few-shot learning methods, graph neural networks and prototypical networks, show that our method can improve the performance of existing models substantially with very little computational overhead, demonstrating the effectiveness of the large margin principle and the potential of our method.

\end{abstract}

\section{Introduction}

Few-shot learning \cite{Feifei06} is a very challenging problem as it aims to learn from very few labeled examples. Due to data scarcity, training a conventional end-to-end supervised model such as deep learning models \cite{Krizhevsky12,He16} from scratch will easily lead to overfitting, and techniques such as data augmentation and regularization cannot solve this problem.

One successful perspective for tackling few-shot learning is meta-learning. Unlike traditional supervised learning that requires a large labeled set for training, meta-learning trains a classifier that can generalize to new tasks by distilling knowledge from a large number of similar tasks and then transferring the knowledge to quickly adapt to new tasks. Several directions have been explored for meta learning, including learn to fine-tune \cite{Ravi17,Finn17,Munkhdalai17,Li17}, sequence based methods \cite{Santoro16,Mishra18}, and metric based learning \cite{Vinyals16,Koch15}. 

Metric based few-shot learning has attracted a lot of interest recently \cite{Vinyals16,Snell17,Fort17,Sung18,Garcia18,Mehrotra17,Koch15,Ren18}, probably due to its simplicity and effectiveness. The basic idea is to learn a metric which can map similar samples close and dissimilar ones distant in the metric space so that a query can be easily classified. Various metric based methods such as siamese networks \cite{Koch15}, matching networks \cite{Vinyals16}, prototypical networks \cite{Snell17}, and graph neural networks \cite{Garcia18} differ in their ways of learning the metric. 

The success of metric based methods relies on learning a discriminative metric space. However, due to data scarcity in the training tasks, it is difficult to learn a good metric space. To reach the full potential of metric based few-shot learning, we propose a large margin principle for learning a more discriminative metric space. The key insight is that samples from different classes should be mapped as far apart as possible in the metric space to improve generalization and prevent overfitting. The large margin constraint has not been enforced in existing metric based methods.

To fill this gap, we develop a unified framework to impose the large margin constraint. In particular, we augment the linear classification loss function of a metric learning method with a distance loss function -- the triplet loss \cite{Schroff15} to train a more metric space. Our framework is simple, robust, very easy to implement, and can be potentially applied to many metric learning methods that adopts a linear classifier. Applications on two state-of-the-art metric learning methods -- graph neural networks \cite{Garcia18} and prototypical networks \cite{Snell17} show that the large margin constraint can substantially improve the generalization capacity of the original models with little computational overhead. Besides the triplet loss, we also explore other loss functions to enforce the large margin constraint. All experimental results confirm the effectiveness of the large margin principle. 

\begin{figure*}[t!]
	\centering
    \includegraphics[width=1\textwidth]{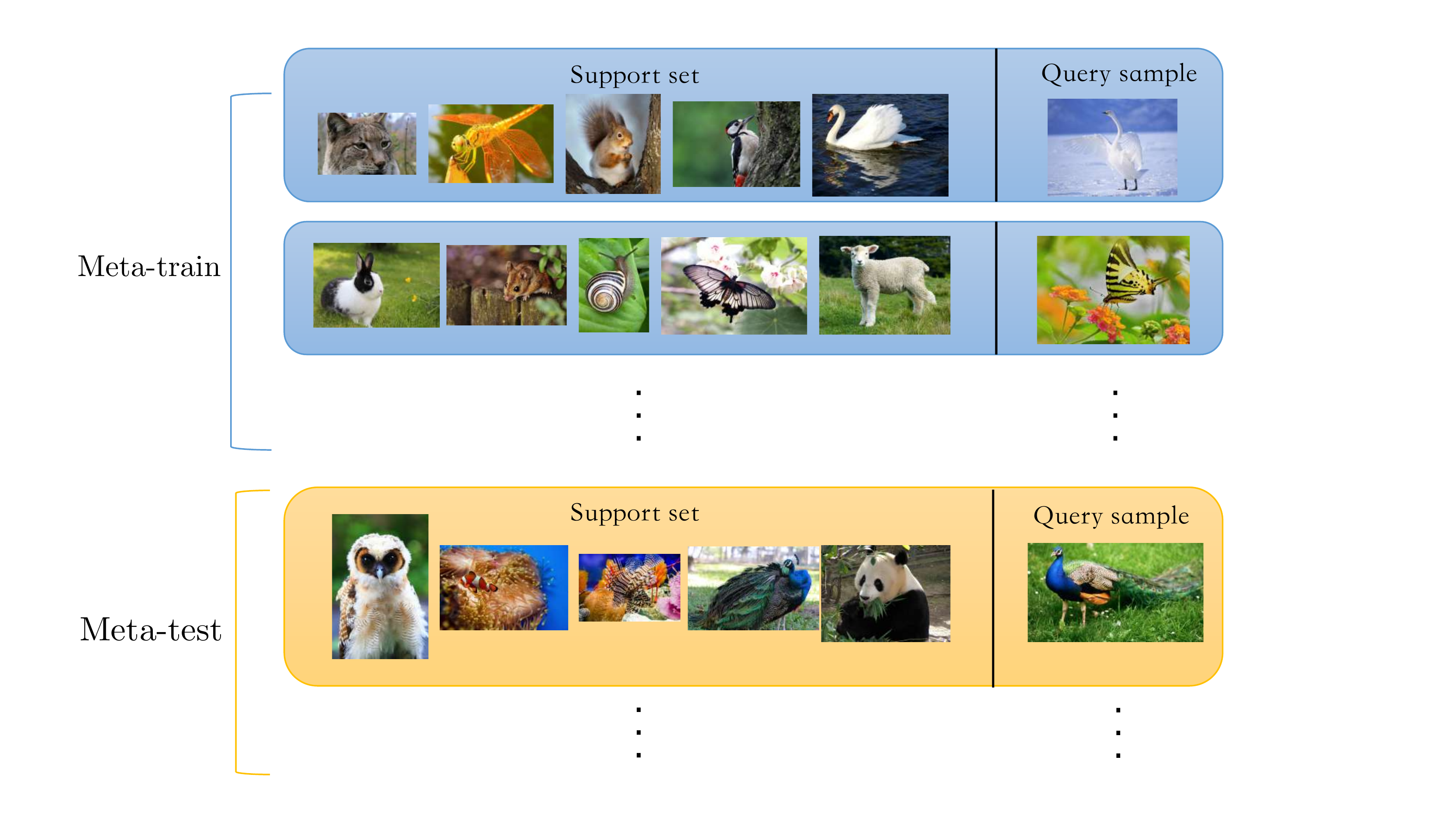}
	\caption{Training and testing process of few-shot learning.}
	\label{fig:image_compressed}
\end{figure*}

Although large margin methods have been widely studied in many areas of machine learning, this paper is the first to investigate its applicability and usefulness in few-shot learning (meta-learning), to our best knowledge. It should be noted that the few-shot learning problem considered here has very different setup with the attributed-based few-shot \cite{LiXin15} or zero-shot learning \cite{Lampert09,Akata13,FuYanWei15,Xian18}. The contributions of this paper include 1) proposing a large margin principle to improve metric based few-shot learning, 2) developing an effective and efficient framework for large margin few-shot learning, and 3) conducting extensive experiments to validate our proposals.

\section{Large Margin Few-Shot Learning} \label{sec:our method}

\subsection{Few-Shot Learning}
Few-shot learning aims to train a classifier which can quickly adapt to new classes and learn from only a few examples. It consists of two phases: meta-training and meta-testing (Fig. \ref{fig:image_compressed}). In meta-training, a large amount of training data $\mathcal{D}_{\text{meta-train}}= \{(\mathbf{x}_i, y_i)\}_{i = 1}^N$ from a set of classes $\mathcal{C}_{\text{train}}$ are used for training a classifier, where $\mathbf{x}_i$ is the feature vector of an example, $y_i\in \mathcal{C}_{\text{train}}$ is the label, and $N$ is the number of training examples. In meta-testing, a support set of $S$ labeled examples $\mathcal{D}_s=\{(\mathbf{x}_j, y_j)\}_{j=1}^{S}$ from a set of new classes $\mathcal{C}_{\text{test}}$ is given, i.e., $y_j\in \mathcal{C}_{\text{test}}$ and $\mathcal{C}_{\text{train}}\cap \mathcal{C}_{\text{test}}=\emptyset$. The goal is to predict the labels of a query set $\mathcal{D}_q=\{(\mathbf{x}_j)\}_{j=S+1}^{S+Q}$, where $Q$ is the number of queries. If the support set $\mathcal{D}_s$ contains $K$ examples from each of $C$ classes, i.e., $S=K\times C$, the few-shot learning problem is called $C$-way $K$-shot learning. Typically, $K$ is a small number such as 1 or 5.

To improve generalization, an episode-based training strategy \cite{Vinyals16} is commonly adopted to better exploit the training set $\mathcal{D}_{\text{meta-train}}$. In particular, in meta-training, a model is trained by support/query sets, where a support set is formed by sampling $K$ examples from each of $C$ classes from $\mathcal{C}_{train}$, and a query set is formed by sampling from the rest of the $C$ classes' samples. The purpose is to mimic the test scenario during training.

\subsection{Large Margin Principle}

\begin{figure}[t!]
	\centering
	\scalebox{1.0}{
		\subfigure[]{
			\includegraphics[width=0.31\textwidth]{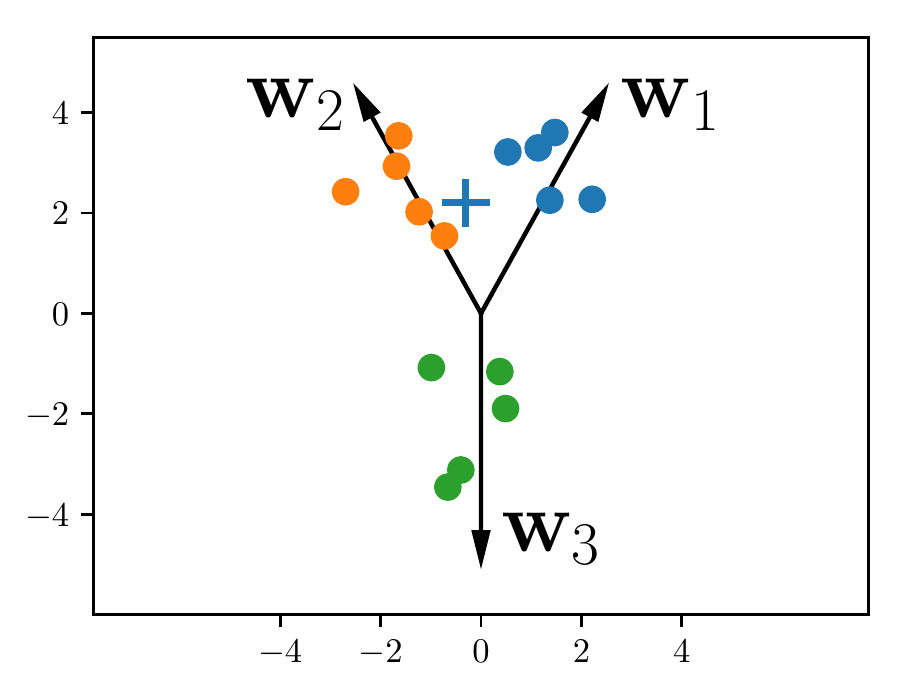}
			\label{fig: non-uniform decision boundary}}
		\subfigure[]{
			\includegraphics[width=0.31\textwidth]{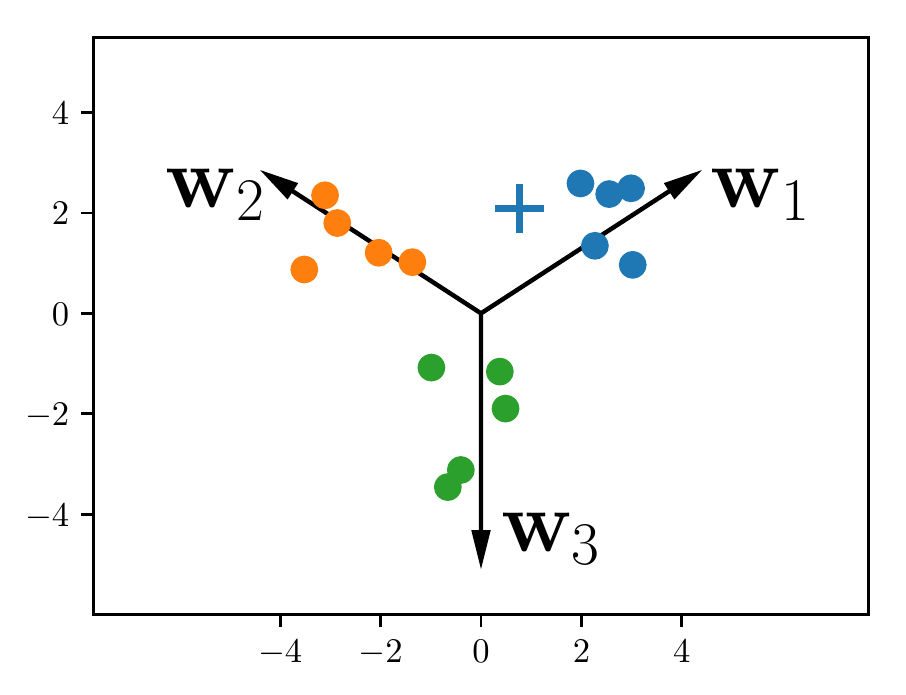}
			\label{fig: uniform decision boundary}}
		\subfigure[]{
			\includegraphics[width=0.31\textwidth]{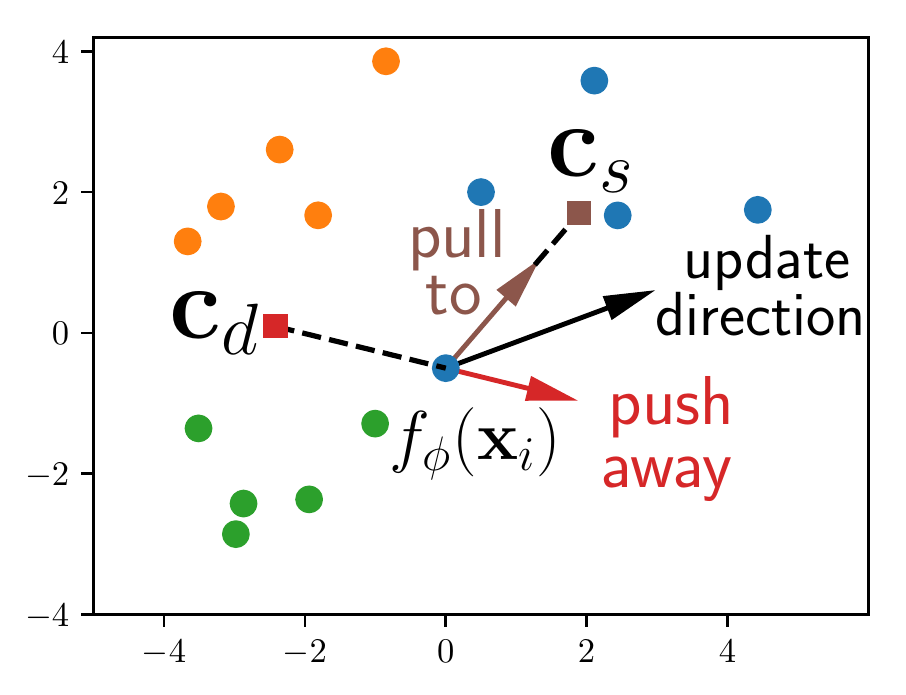}
			\label{fig:gradient}}}
	\caption{Large margin few-shot learning. (a) Classifier trained without the large margin constraint. (b) Classifier trained with the large margin constraint. (c) Gradient of the triplet loss. }
	\label{fig:idea}
\end{figure}

\textbf{How Few-Shot Learning Works.} To learn quickly from only a couple of examples whose classes are unseen in meta-training, a model should acquire some transferable knowledge in meta-training. In metric-based few-shot learning \cite{Koch15,Garcia18,Snell17}, the basic idea is to learn a nonlinear mapping $f_\phi(\cdot)$ which can model the class relationship among data samples, i.e., similar samples are mapped to nearby points in the metric space while dissimilar ones are mapped far apart. Usually, the mapping $f_\phi(\cdot)$ embeds a sample $\mathbf{x}_i$ into a relatively low dimensional space, and the embedded point $f_\phi(\mathbf{x}_i)$ is then classified by a linear classifier, e.g., the softmax classifier. Note that the softmax classifier can be considered as the last fully connected layer in a neural network. Both the mapping $f_\phi(\cdot)$ and the classifier parameters are learned by minimizing the cross entropy loss:
\begin{equation}\label{eq:softmax_loss}
\mathcal{L}_{\text{softmax}}=\frac{1}{N}\sum _i-\log(\frac{e^{\mathbf{w}_{y_i}^\top f_\phi(\mathbf{x}_i)}}{\sum _j e^{\mathbf{w}_{j}^\top f_\phi(\mathbf{x}_i)}}),
\end{equation}
where $\mathbf{w}_j$ is a classifier weight vector corresponding to the $j$-th column of the weight matrix $W$ of the softmax classifier. Without loss of generality, we omit the bias $b$ to simplify the analysis. Note that $\mathbf{w}_j$ can be considered as the class center of samples in class $j$ in the embedding space.

After learning $f_\phi(\cdot)$ and $W$, the model can be used for testing. Fig. \ref{fig: non-uniform decision boundary} shows a 3-way 5-shot test case, where the support samples are indicated by dots and the query sample is indicated by a cross. Samples in the same class are indicated by the same color. We can see that samples from each class are mapped to cluster around the corresponding classifier weight vector $\mathbf{w}_j$. However, the query sample, which belongs to class 1, may be wrongly classified to class 2, due to the small margin between  $\mathbf{w}_1$ and $\mathbf{w}_2$.

\textbf{How Can It Work Better?} As each training episode consists very few samples of each class, the standard error of the sample mean is high \cite{Mood74}. In other words, the class average may be a poor estimate of the true class center, and some samples may not well represent its own class. Hence, the model may not be able to learn a discriminative metric space. 

To alleviate this problem and improve the model generalization capacity on new classes, we propose to enforce a large margin between the classifier weight vectors (or class centers). The idea is that samples from different classes should be mapped as far apart as possible in the metric space. As shown in Fig. \ref{fig: uniform decision boundary}, the query sample can be correctly classified by enlarging the margin between $\mathbf{w}_1$ and $\mathbf{w}_2$. It is worth noting that the large margin principle makes the classifier weight vectors distributed in a balanced manner (Fig. \ref{fig: uniform decision boundary}), which leads to balanced decision boundaries.

\subsection{Model}

To enforce the large margin constraint, we propose to augment a large margin loss function to the classification loss function, and the total loss is given by
\begin{equation}\label{eq:total_loss}
\begin{split}
\mathcal{L} =\mathcal{L}_{\text{softmax}} + \lambda*\mathcal{L}_{\text{large-margin}},
\end{split}
\end{equation}
where $\lambda$ is a balancing parameter. We choose the triplet loss \cite{Schroff15} to be the large margin distance function, which acts on the embeddings of training samples in the metric space:
\begin{equation}\label{eq:triplet_loss}
\begin{split}
\mathcal{L}_{\text{large-margin}}=\frac{1}{N_t}\sum_{i=1}^{N_t}\left[\parallel f_\phi(\mathbf{x}_i^a)-f_\phi(\mathbf{x}_i^p)\parallel_2^2- \parallel f_\phi(\mathbf{x}_i^a)-f_\phi(\mathbf{x}_i^n)\parallel_2^2+m\right]_+,
\end{split}
\end{equation}
where $m \in \mathbb{R}_{++} $ is a parameter for margin. $(\mathbf{x}_i^a, \mathbf{x}_i^p, \mathbf{x}_i^n)$ forms a triplet, where $\mathbf{x}_i^a$ is called the anchor sample, $\mathbf{x}_i^p$ is the positive sample w.r.t. the anchor sample, and $\mathbf{x}_i^n$ is the negative sample w.r.t. the anchor sample. $\mathbf{x}_i^a$ and $\mathbf{x}_i^p$ have same labels, and $\mathbf{x}_i^a$ and $\mathbf{x}_i^n$ have different labels. $N_t$ is the number of triplets. Any sample in the training set can be chosen as anchor. Once selected, a plurality of positive samples and negative samples are paired with the anchor sample to form a plurality of triplets.

Intuitively, the augmented triplet loss will help train a more discriminative mapping $f_\phi(\cdot)$ which embeds samples of the same class closer and those of different classes farther in the metric space. Next we provide a theoretical analysis to show how the triplet loss reshapes the embeddings.

\subsection{Analysis}

We analyze the influence of the augmented triplet loss on the embeddings by studying the gradient of $\mathcal{L}_{\text{large-margin}}$ with respect to $f_\phi(\mathbf{x}_i)$, the embedding of sample $\mathbf{x}_i$, during back propagation.

Since terms with zero loss in \eqref{eq:triplet_loss} have no effect on the gradients, we only need to consider the triplets whose loss within the square brackets is positive and so the hinge operation $[\cdot]_+$ can be removed. To find
the gradient of $\mathcal{L}_{\text{large-margin}}$ with respect to $f_\phi(\mathbf{x}_i)$, we need to find the terms in which $\mathbf{x}_i$ is the anchor sample, or the positive sample, or the negative sample. We partition the samples into three multisets. The first set $S_s$ contains all samples that are paired with with $\mathbf{x}_i$ and have the same label as $\mathbf{x}_i$. The second set $S_d$ contains all samples that are paired with with $\mathbf{x}_i$ but have different label with $\mathbf{x}_i$. The third set contains samples that are not paired with $\mathbf{x}_i$. The multiplicity of a element in these multisets is the number of triplets in which the element is paired with $\mathbf{x}_i$.
Note that if a sample $\mathbf{x}_s\in S_s$, the distance $\parallel{f_\phi(\mathbf{x}_i)-f_\phi(\mathbf{x}_s)}\parallel^2_2$ is added to the loss, while if a sample $\mathbf{x}_d\in S_d$, the distance is subtracted from the loss. After some rearrangements, \eqref{eq:triplet_loss} can be written as
\begin{equation}
\begin{split}
	\mathcal{L}_{\text{large-margin}} = \frac{1}{N_t}\left(\sum_{\mathbf{x}_s  \in S_s}\parallel{f_\phi(\mathbf{x}_i)-f_\phi(\mathbf{x}_s)}\parallel^2_2 -\sum_{\mathbf{x}_d \in S_d}\parallel{f_\phi(\mathbf{x}_i)-f_\phi(\mathbf{x}_d)}\parallel^2_2\right) + \text{const},
\end{split}
\end{equation}
where const denotes a constant independent of $\mathbf{x}_i$. Then the gradient of $\mathcal{L}_{\text{large-margin}}$ with respect to $f_\phi(\mathbf{x}_i)$ can be derived as:
\begin{align}
	\frac{\partial\mathcal{L}_{\text{large-margin}}}{\partial f_\phi(\mathbf{x}_i)}
	&= - \frac{2|S_s|}{N_t}
			\left(\frac1{|S_s|}\sum_{\mathbf{x}_s  \in S_s}f_\phi(\mathbf{x}_s)-f_\phi(\mathbf{x}_i)\right)
		-  \frac{2|S_d|}{N_t}
			\left(f_\phi(\mathbf{x}_i)-\frac1{|S_d|}\sum_{\mathbf{x}_d\in S_d}f_\phi(\mathbf{x}_d)\right) \\
	&= - \underbrace{\frac{2|S_s|}{N_t}
			\left(\mathbf{c}_s-f_\phi(\mathbf{x}_i)\right)
		}_\text{pull to its own class}
		- \underbrace{\frac{2|S_d|}{N_t}
			\left(f_\phi(\mathbf{x}_i)-\mathbf{c}_d\right)
		}_\text{push away from other classes} \label{eq:gradient}
\end{align}
where $\mathbf{c}_s$ is the center of embedded points in $S_s$, and $\mathbf{c}_d$ is the center of embedded points in $S_d$. By \eqref{eq:gradient}, the gradient consists of two parts. The first part is a vector pointing from $f_\phi(\mathbf{x}_i)$ to the center of embedded points in $S_s$, which pulls $f_\phi(\mathbf{x}_i)$ to its own class during training, as indicated by the brown arrow in Fig.
 \ref{fig:gradient}.
The other part is a vector pointing from the center of embedded points in $S_d$ to $f_\phi(\mathbf{x}_i)$, which pushes $f_\phi(\mathbf{x}_i)$ away from other classes, as indicated by the red arrow in Fig. \ref{fig:gradient}. This shows that the augmented triplet loss can effectively enforce the large margin constraint.

\subsection{Discussion}

\textbf{When Does It Work?} The working assumption of the triplet loss is that the similarity/dissimilarity between the embedded points can be measured by Euclidean distance. If the embedded points lie on a nonlinear manifold in the metric space, Euclidean distance cannot reflect their similarity. This indicates that the embedded points should be linearly separable in the metric space, which suggests that the triplet loss should work with a linear model such as the softmax classifier in the metric space.

\textbf{Computational Overhead.} The computational overhead of the augmented triplet loss lies in two aspects: triplet selection and loss computation. It is well known that online selection of triplets could be very time-consuming when the training set is large. However, in few-shot learning, the number of samples for each class is fixed in each update iteration, so we can use an offline strategy for triplet selection. In fact, we only need to form the triplets once, and then store the parings and use them for indexing the embeddings in each update. In this way, the computational overhead for triplet selection is negligible. For the loss computation, the time complexity for each update is $\mathcal{O}(N_td)$ where $N_t$ is the number of triplets and $d$ is the dimensionality of embeddings. Hence, the computation is efficient if the embeddings are low-dimensional (see Section \ref{sec:GNN-experiment} and  \ref{sec:PN-experiment} for the report of running times).

\textbf{Alternative Loss Functions.} Besides the triplet loss, we also explore other loss functions for incorporating the large margin constraint for few-shot learning, including the normalized triplet loss, the normalized contrastive loss \cite{Hadsell06,Sun14}, the normface loss \cite{Wang17}, the cosface loss \cite{WangHao18}, and the arcface loss \cite{Deng18}. We discuss and compare these methods in the section \ref{sec:discussion}.

\section{Large Margin Graph Neural Networks} \label{sec:GNN}

In this section, we apply the proposed large margin framework on the recently proposed graph neural networks (GNN) \cite{Garcia18} for few-shot learning, which achieved state-of-the-art performance on some benchmark datasets for both few-shot learning and semi-supervised few-shot learning. 

\subsection{Graph Neural Networks}

In the training of GNN, each mini-batch consists of multiple episodes, and the number of episodes is the batch size. For each episode, the meta-train samples are given as:
\begin{equation}
\begin{split}
\mathcal{D}_{\text{meta-train}}=\{\{(\mathbf{x}_1,y_1),...(\mathbf{x}_S,y_S)\}, \{\overline{\mathbf{x}}_{S+1},\overline{y}_{S+1}\},
\{\tilde{\mathbf{x}}_1,\tilde{\mathbf{x}}_2,...,\tilde{\mathbf{x}}_r\}:y_i, \overline{y}_{S+1}\in\{1,...,C\}\},
\end{split}
\end{equation}
where $\{(\mathbf{x}_1,y_1),...(\mathbf{x}_S,y_S)\}$ is the support set with labels and $\{\overline{\mathbf{x}}_{S+1},\overline{y}_{S+1}\}$ is the query set with only one query. As we also consider the semi-supervised setting for few-shot learning, $\{\tilde{\mathbf{x}}_1, \tilde{\mathbf{x}}_2,...,\tilde{\mathbf{x}}_r\}$ is the set of samples without labels. For few-shot learning, $r$ is $0$; and for semi-supervised few-shot learning, $r\ge 1$.
For each sample, the initial feature vector is given by:
$$\mathbf{x}_i^{(0)} = (\theta(\mathbf{x}_i); h(y_i)),$$ 
where $\theta$ is a trainable convolutional neural network, and $h(\cdot)$ is a one-hot vector encoding the label information. Denote by
$\mathbf{X}^{(0)}=[\mathbf{x}_1^{(0)},...,\mathbf{x}_{S}^{(0)},\overline{\mathbf{x}}_{S+1}^{(0)},\tilde{\mathbf{x}}_{1}^{(0)},...,\tilde{\mathbf{x}}_{r}^{(0)}]^\top$ the initial feature matrix.

A graph is constructed by taking each sample as a node, and the adjacency matrix $\tilde{A}^{(0)}$ is updated by:
$$\tilde{A}_{i,j}^{(0)}=\textrm{MLP}_{\tilde{\theta}}(\textrm{abs}(\mathbf{x}_i^{(0)}-\mathbf{x}_j^{(0)})),$$
where $\textrm{MLP}_{\tilde{\theta}}(\cdot)$ is a multilayer perceptron with learnable parameters $\tilde{\theta}$ and $\textrm{abs}(\cdot)$ is the absolute value function.
The operator family is formed by $\mathcal{A}=\{\tilde{A}^{(0)},\mathbf{1}\}$, where $\mathbf{1}$ is an all-$1$'s matrix.
The new features will be obtained by combining the features of its adjacent nodes and then projected by a projection layer:
$$\mathbf{X}^{(1)}=\sigma(\sum_{B\in \mathcal{A}}B\mathbf{X}^{(0)}\theta_{B}^{(0)}),$$
where $\theta_{B}^{(0)}\in \mathbb{R}^{d_0\times d_1}$ is the learnable parameters of the projection layer, $d_0$ is the length of $\mathbf{x}_i^{(0)}$, and $d_1$ is the length of $\mathbf{x}_i^{(1)}$. This process can be repeated for several times. After $M$ iterations, we obtain the final embedding matrix $\mathbf{X}^{(M)}$. 

To classify the embeddings, GNN uses a parametric softmax classifier. Namely, for $C$-way learning, in each training episode, the probability of the query being classified as the $k$-th class is:
$$P(k|\overline{\mathbf{x}}_{S+1})=\frac{\exp(\mathbf{w}_k^\top f_\phi(\overline{\mathbf{x}}_{S+1}))}{\sum_{j=1}^C \exp(\mathbf{w}_j^\top f_\phi(\overline{\mathbf{x}}_{S+1}))},$$
where $f_\phi(\overline{\mathbf{x}}_{S+1})=\overline{\mathbf{x}}_{S+1}^{(M)}$.
For all episodes in a mini-batch, the softmax loss is:
$$\mathcal{L}_{\text{softmax}}=-\sum_k \overline{y}_{S+1,k}\log P(Y_* = \overline{y}_{S+1,k} | \overline{\mathbf{x}}_{S+1,k}),$$
where $Y_*$ is the predicted label for the query in the $k$-th episode, $\overline{\mathbf{x}}_{S+1,k}$ is the feature of the query, and $\overline{y}_{S+1,k}$ is the ground truth label.

\subsection{Large Margin Graph Neural Networks}
GNN models the class relationship of input samples using graph operators, and maps samples with the same label representation -- the one hot vector $h(y_i)$ to a fixed weight vector $\mathbf{w}_{y_i}$ in the embedding space. To make GNN more discriminative, we augment a triplet loss on its objective function to train a large margin graph neural network (L-GNN). The triplet loss is defined as in \eqref{eq:triplet_loss},
where $f_\phi(\mathbf{x}_{i})=\mathbf{x}_{i}^{(M)}$ is the embedding of $\mathbf{x}_{i}$ by GNN.

\textbf{Triplet Selection.} In the training of GNN, each mini-batch consists of multiple episodes. For $C$-way $K$-shot learning, the support set of each episode is formed by sampling $K$ examples from each of the $C$ classes, resulting in $C\times K$ examples. The label representation of each example is a $C$-dimensional one-hot vector. For any two examples, if their label representations are the same, they should be mapped to the same class, and vice versa. To form the triplets, for each example (anchor) in the support set, we sample $5$ positive examples from the training batch which have the same label representation with the anchor; and for each positive example, we sample $5$ negative examples which have different label representation with the anchor. Hence, for each anchor example, $25$ triplets are formed. For $5$-way $5$-shot learning, with a batch size of $40$, there will be a total of $25,000\ (25\times25\times 40)$ triplets. In our experiments on Mini-Imagenet, it only takes $0.331$s to form the triplets. Since the selection of triplets only needs to be done once at the beginning of training, it incurs almost no computational overhead. 

\subsection{Experimental Results}\label{sec:GNN-experiment}

\begin{figure*}[]
	\centering
	\subfigure[Samples of Omniglot.]{
		\includegraphics[width=0.36\textwidth]{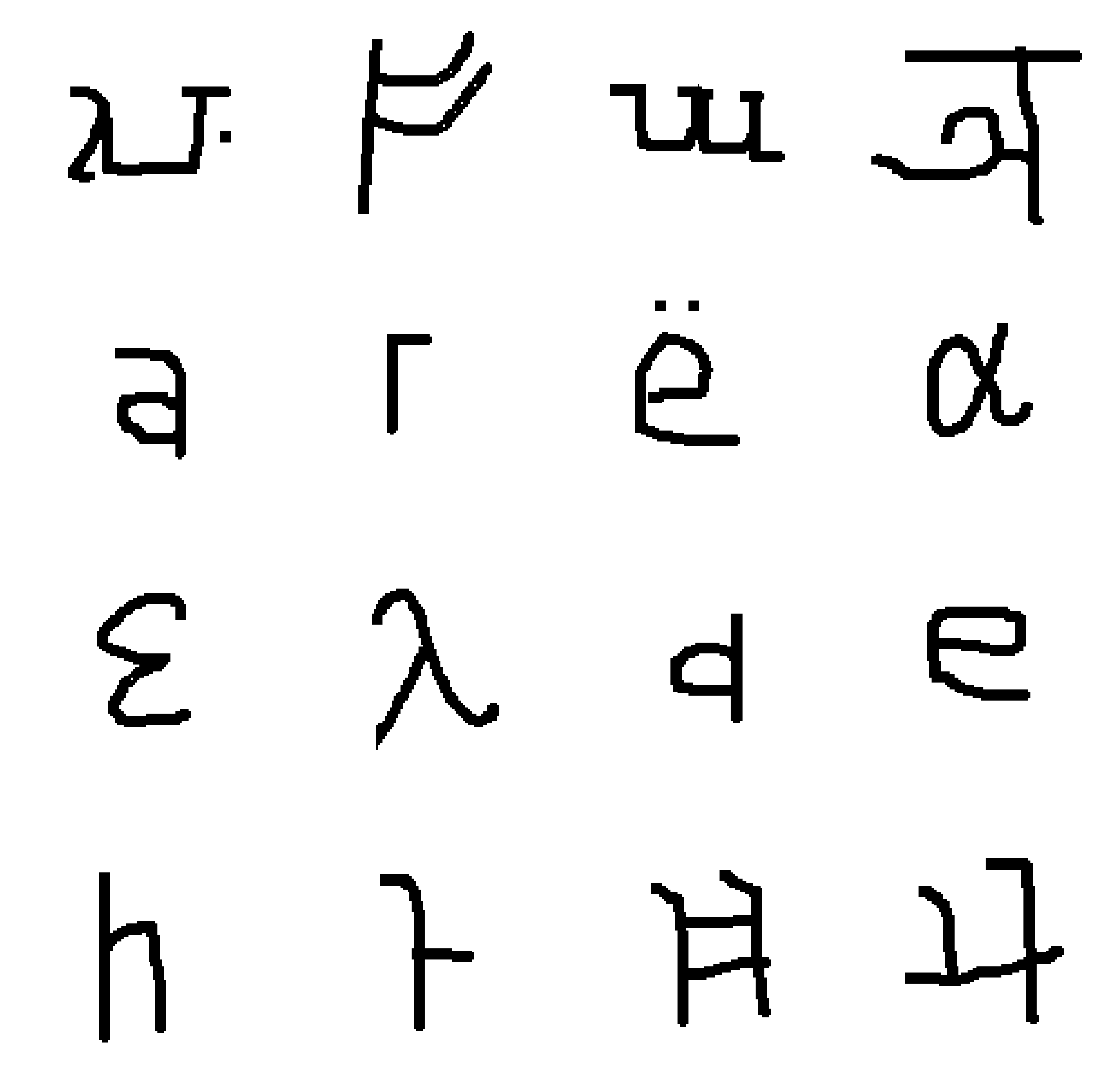}}
	\subfigure[Samples of Mini-Imagenet.]{
		\includegraphics[width=0.5\textwidth]{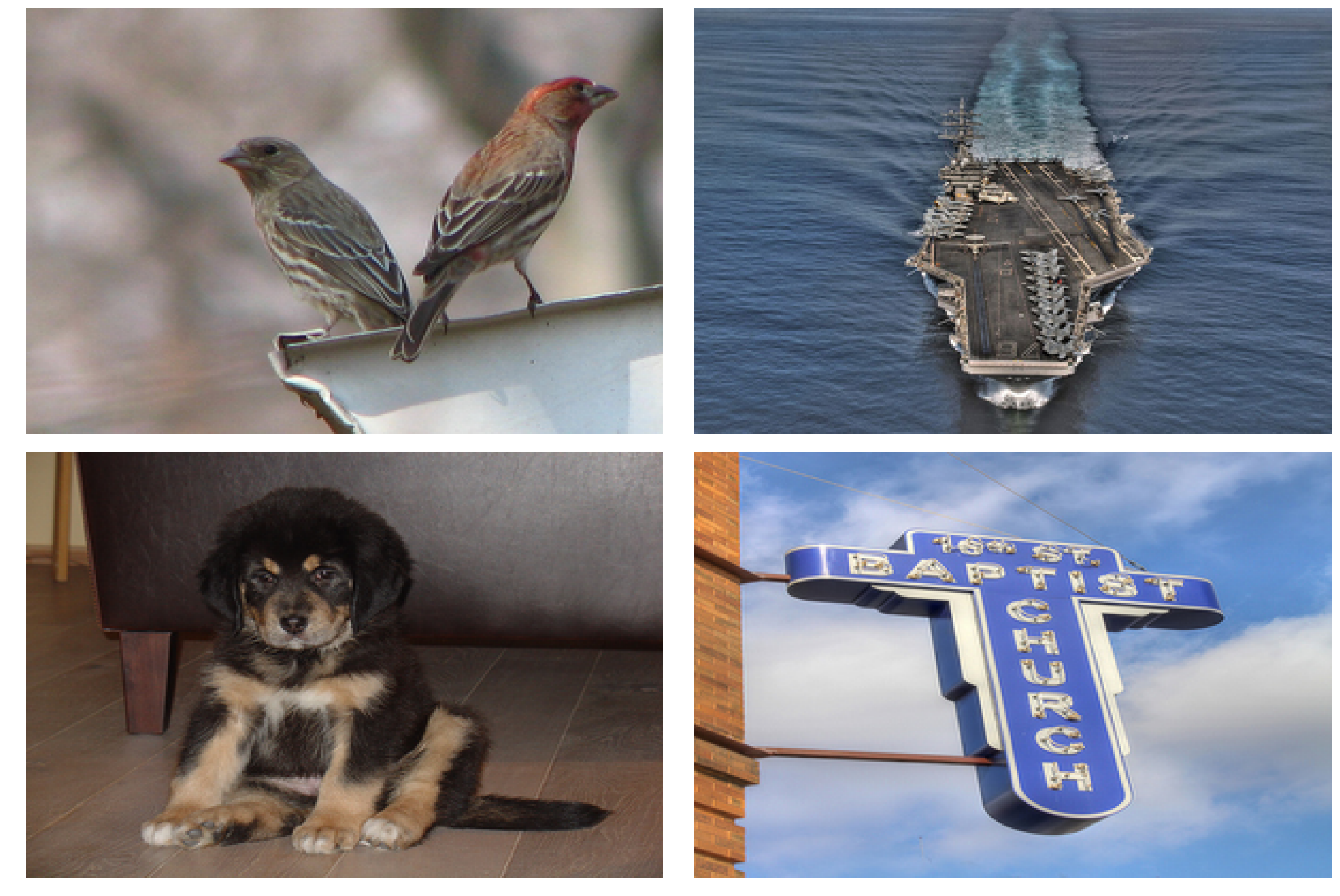}}
	\caption{Image samples of Omniglot and Mini-Imagenet.}
	\label{fig: dataset}
\end{figure*}

\textbf{Datasets.} The experiments are conducted on two benchmark datasets: Omniglot and Mini-Imagenet. The Omniglot dataset consists of $1,623$ characters from $50$ different alphabets drawn by different individuals. Each character has $20$ samples and each sample is rotated by $90^\circ$, $180^\circ$, $270^\circ$ to enlarge the dataset by four times. The dataset is split into $1,200$ characters plus rotations for training and $423$ characters plus rotations for testing. On Omniglot, we consider $5$-way and $20$-way for both $1$-shot and $5$-shot learning. The Mini-Imagenet dataset is composed of $60,000$ images from the Imagenet dataset, and it has $100$ classes with $600$ samples from each class. It is split into $64$, $16$, and $20$ disjoint classes for training, validation, and testing respectively. On Mini-Imagenet, we consider $5$-way and $10$-way for both $1$-shot and $5$-shot learning. Fig. \ref{fig: dataset} shows some image samples of Omniglot and Mini-Imagenet.

\textbf{Parameter Setup.} To make sure our method can work well in practice, we use fixed parameters in all experiments. We set the balancing parameter $\lambda=1$ in \eqref{eq:total_loss}. For the triplet loss, we set the margin as $\frac{1}{2N_b}\sum_{i=1}^{N_b} \parallel f_\phi(\mathbf{x}_i)\parallel _2$, the average of the $L_2$-norm of all embeddings in one mini-batch with randomly initialized model parameters at the beginning of training, where $N_b$ is the number of all samples in one mini-batch.

\begin{table*}[t]
\center
\resizebox{0.9\textwidth}{!}{
\begin{tabular}{c|ccccc}
\toprule

Dataset&Model&1-shot&5-shot&1-shot&5-shot\\
\midrule
&  &\multicolumn{2}{c}{5-Way}&\multicolumn{2}{c}{20-Way}\\
\cmidrule[0.01pt]{2-6}
\multirow{2}{*}{Omniglot}&GNN &$99.20\%$&$99.70\%$&$97.40\%$&$99.00\%$\\
&L-GNN &$99.18\%$ &$\mathbf{99.73\%}$ &$\mathbf{97.55\%}$ &$\mathbf{99.17\%}$ \\
\midrule
 & &\multicolumn{2}{c}{5-Way}&\multicolumn{2}{c}{10-Way}\\
\cmidrule[0.01pt]{2-6}
\multirow{2}{*}{Miniimagenet} &GNN &$50.33\pm 0.36 \%$&$66.41\pm 0.63\%$&$33.16\pm 0.65 \%$&$49.14\pm 0.68\%$\\
&L-GNN &$\mathbf{51.08\pm 0.69\%}$&$\mathbf{67.57\pm 0.66\%}$ & $\mathbf{34.53\pm 0.64\%}$&$\mathbf{51.48\pm 0.69\%}$\\
\bottomrule
\end{tabular}
}
\caption{Few-shot learning on Omniglot and Mini-Imagenet with GNN and L-GNN. Results are averaged with $95\%$ confidence intervals. '$-$': not reported.}
\label{table: accuracy on GNN}
\end{table*}

\begin{table*}[t]
\center
\resizebox{0.9\textwidth}{!}{
\begin{tabular}{cccccc}
\toprule
&&\multicolumn{2}{c}{Omniglot}&\multicolumn{2}{c}{Mini-Imagenet}\\
&Model&$20\%$-labeled&$40\%$-labeled&$20\%$-labeled&$40\%$-labeled\\
\midrule
Trained only with  &GNN  & $99.20\%$ &$99.59\%$& $50.33\pm 0.36\%$ &$56.91\pm 0.42\%$\\
labeled samples&L-GNN &$99.18\%$&$\mathbf{99.65\%}$&$\mathbf{51.08\pm 0.69\%}$&$\mathbf{57.90\pm 0.68}\%$\\
\midrule
\multirow{2}{*}{Semi-supervised}&GNN & $99.59\%$ &$99.63\%$& $52.45\pm 0.88\%$ &$58.76\pm 0.86\%$\\
&L-GNN &$\mathbf{99.70\%}$&$\mathbf{99.71\%}$&$\mathbf{54.51\pm 0.69}\%$&$\mathbf{60.47\pm 0.67}\%$\\
\bottomrule
\end{tabular}
}
\caption{$5$-way $5$-shot semi-supervised few-shot learning on Omniglot and Mini-Imagenet. ``Trained only with labeled samples'' means that the unlabeled samples are not used in training and testing; ``Semi-supervised'' means that the unlabeled samples are used in training and testing. Results are averaged with $95\%$ confidence intervals.}
\label{table: accuracy on Semi-Supervised Learning}
\end{table*}

\textbf{Results on Few-Shot Learning.} The results on the two benchmark datasets are shown in Tabel \ref{table: accuracy on GNN}. On Omniglot, we can see that although GNN has already achieved very high accuracy, our method L-GNN can still improvement further, especially for the more challenging $20$-way classification tasks. On Mini-Imagenet, L-GNN improves GNN on \emph{every} learning task, especially for the more difficult $10$-way classification tasks where the largest improvement is 2.3\% in terms of absolute accuracy. The results clearly demonstrate the benefit of incorporating the large margin loss in training.

\textbf{Results on Semi-Supervised Few-Shot Learning.} Table \ref{table: accuracy on Semi-Supervised Learning} shows the results of semi-supervised 5-way 5-shot learning on Omniglot and Mini-Imagenet. We can see that our method L-GNN consistently outperforms GNN on \emph{all} semi-supervised classification tasks. The improvements are significant on Mini-Imagenet. This shows that adding large margin constraint helps learn a better embedding space for both labeled and unlabeled data, and again demonstrates the effectiveness of our method.

\begin{figure}[!t]
	\centering
	\subfigure[Test accuracy as $m$ changes.]{
		\includegraphics[width=0.8\textwidth]{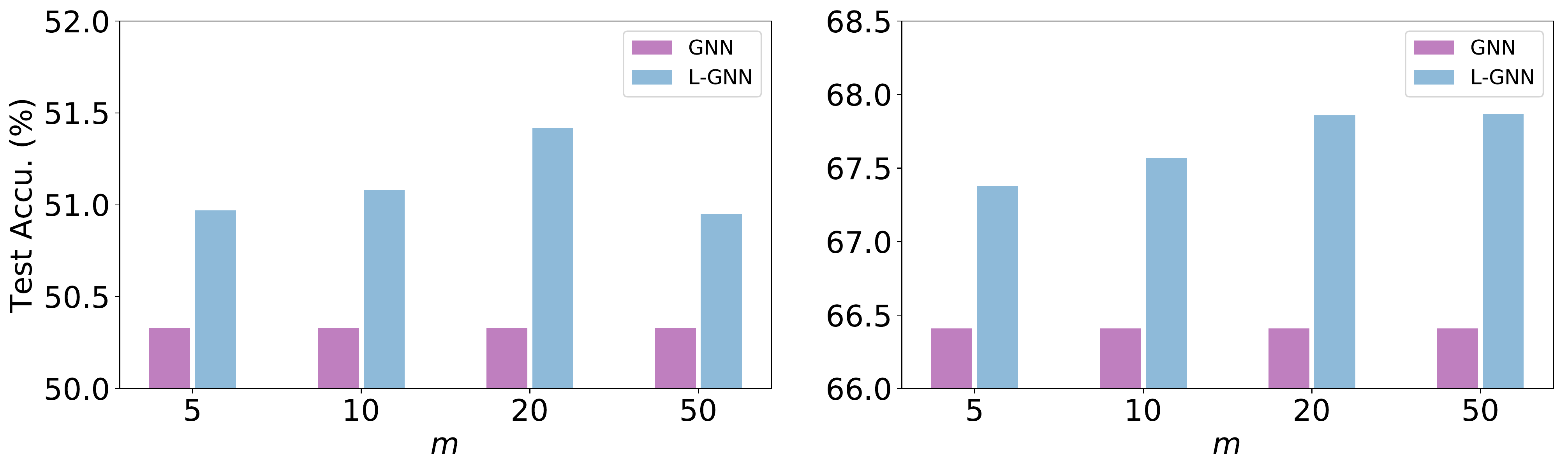}
		\label{fig: gnn-margin}}
	\subfigure[Test accuracy as $\lambda$ changes.]{
		\includegraphics[width=0.8\textwidth]{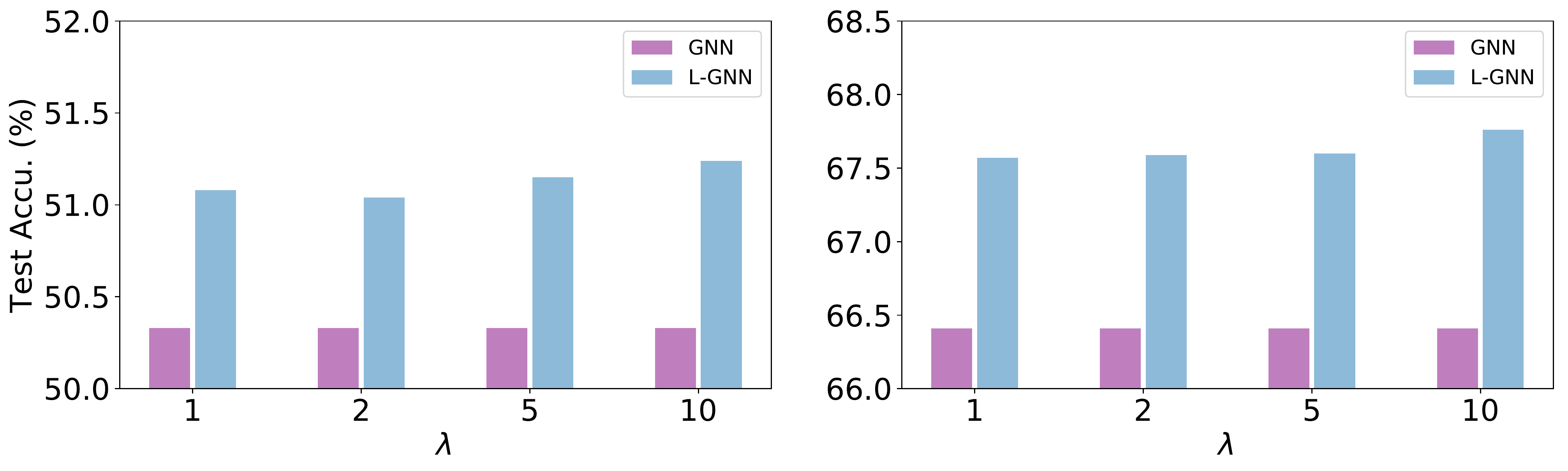}
		\label{fig: gnn-lambda}}
    \captionsetup{justification=centering}

	\caption{$5$-way learning using GNN on Mini-Imagenet. (Left: $1$-shot, right: $5$-shot)}
	\label{fig: robustness gnn}
\end{figure}

\textbf{Parameter Sensitivity.} We also study the sensitivity of the balancing parameter $\lambda$ and the margin $m$ in L-GNN. The experimental results for $5$-way $1$-shot and $5$-shot learning are shown in Fig. \ref{fig: robustness gnn}. Our method L-GNN consistently improves GNN in all cases for a wide range of $\lambda$ and $m$, demonstrating its robustness. 

\textbf{Running Time.} The computational overhead of our method is very small. We evaluate the running time on the platform of Intel(R) Xeon(R) CPU E5-2640 v4, 2.40GHz with GeForce GTX 1080 Ti. For $5$-way $5$-shot learning on Mini-Imagenet, one update of L-GNN takes $0.237$s versus $0.228$s of GNN, which only incurs $3.9\%$ computational overhead.

\section{Large Margin Prototypical Networks} \label{sec:PN}

In this section, we apply the proposed large margin framework on the popular prototypical networks (PN) \cite{Snell17} for few-shot learning. PN is very easy to implement and efficient to train, and achieved very competitive performance on some benchmark datasets.

\subsection{Prototypical Networks}
PN is constructed based on the following steps.
A training set with $N$ labeled examples $\mathcal{D}_{\text{meta-train}}=\{(\mathbf{x}_1, y_1),...,(\mathbf{x}_N, y_N)\}, y_i\in\{1,...,C_{\text{train}}\}$ is given. 
First, randomly sample $N_c$ classes from $\{1,...,C_{\text{train}}\}$, and denote by $C_{N_c}$ the set of class indices. Denote by $c_k$ the $k$-th element in $C_{N_c}$ and by $\mathcal{D}_{c_k}$ the subset of all training samples with $y_i=c_k$.
For each class in $C_{N_c}$, randomly select some samples from $\mathcal{D}_{c_k}$ to form $\mathcal{S}_k$ which is a subset of the support set; and randomly select some other samples from $\mathcal{D}_{c_k}$ to form $\mathcal{Q}_k$ which is a subset of the query set, and make sure $\mathcal{S}_k\cap \mathcal{Q}_k=\emptyset$. Then for each class, compute the prototype $\mathbf{c}_k=\frac{1}{\left| \mathcal{S}_k\right|}\sum _{(\mathbf{x}_i,y_i)\in \mathcal{S}_k}f_\phi(\mathbf{x}_i)$ where the mapping $f_\phi(\cdot)$ is typically a convolutional neural network with learnable parameters $\phi$. 

PN uses a non-parametric softmax classifier. Namely, for a query sample $\textbf{x}_i$ in $\mathcal{Q}_k$, the probability of it being classified to the $k$-th class is:
\begin{align}\label{eq:pn-classifier}
P(k|\textbf{x}_i\in\mathcal{Q}_k)=\frac{\exp(-d(f_\phi(\mathbf{x}_i),\mathbf{c}_k))}{\sum_{j=1}^{N_c} \exp(-d(f_\phi(\mathbf{x}_i),\mathbf{c}_j))},
\end{align}
where $d(\cdot,\cdot)$ is a metric measuring the distance between any two vectors, which can be cosine distance or Euclidean distance. 
For all query samples in an episode, the classification loss is:
\begin{align}\label{eq. pn softmax loss}
\mathcal{L}_{\text{softmax}} = -\frac{1}{ N_c}\sum_{k=1} ^{N_c}\frac{1}{\left| \mathcal{Q}_k\right|}\sum_{(\mathbf{x}_i,y_i)\in \mathcal{Q}_k}P(y_i|\textbf{x}_i).
\end{align}

If $d(\cdot,\cdot)$ is Euclidean distance, PN is actually a linear model in the embedding space \cite{Snell17}. Since
\begin{equation}
\begin{split}
-\parallel f_\phi (\mathbf{x})-\mathbf{c}_k\parallel ^2&=-f_\phi (\mathbf{x})^T f_\phi (\mathbf{x})+2\mathbf{c}_k^Tf_\phi (\mathbf{x})-\mathbf{c}_k^T\mathbf{c}_k\\
&=\mathbf{w}_k^Tf_\phi (\mathbf{x})+b_k+\textrm{const}, \nonumber
\end{split}
\end{equation}
where $\mathbf{w}_k=2\mathbf{c}_k$ and $b_k=-\mathbf{c}_k^T\mathbf{c}_k$, \eqref{eq:pn-classifier} can be considered as a linear classifier. 

\subsection{Large Margin Prototypical Networks}

PN models the class relationship between the query and support samples by measuring the distance between the query and the class centers of support samples in the embedding space. To make PN more discriminative, we augment a triplet loss on its objective function to train a large margin prototypical network (L-PN). The triplet loss is defined as in \eqref{eq:triplet_loss},
where $f_\phi(\mathbf{x}_{i})$ is the embedding of input sample $\mathbf{x}_{i}$ by PN.

\textbf{Triplet Selection.} The implementation of prototypical networks does not use mini-batch in an update iteration. Take $5$-shot learning for example, in one update iteration, for each class, the general practice \cite{Snell17} is to sample $5$ support examples and extra $15$ query examples, so the number of samples in each class is $20$. For each sample (anchor) in the support set and the query set, we sample $10$ positive examples from the class of anchor; and for each positive sample, we sample $10$ negative examples from other classes. Hence, for each sample, $100$ triplets are formed. For $5$-way $5$-shot learning, with $N_c=20$ classes, a total of $40,000\ (100\times20\times20)$ triplets are formed. In our experiments on Mini-Imagenet, it only takes $0.118$s to form the triplets. Since the selection of triplets only needs to be done once at the beginning of training, it incurs almost no computational overhead.

\subsection{Experimental Results} \label{sec:PN-experiment}

\begin{table*}[t]
\center
\resizebox{0.9\textwidth}{!}{
\begin{tabular}{c|cccccc}
\toprule

Dataset&Model&Dist&1-shot&5-shot&1-shot&5-shot\\
\midrule
& & &\multicolumn{2}{c}{5-Way}&\multicolumn{2}{c}{20-Way}\\
\cmidrule[0.01pt]{2-7}
\multirow{2}{*}{Omniglot}&PN &Euclid.&$98.49\%$&$99.60\%$&$95.21\%$ &$98.59\%$ \\
&L-PN &Euclid.&$98.46\%$&$\mathbf{99.63\%}$&$95.16\%$ &$\mathbf{98.69\%}$ \\
\midrule
& & &\multicolumn{2}{c}{5-Way}&\multicolumn{2}{c}{10-Way}\\
\cmidrule[0.01pt]{2-7}
\multirow{4}{*}{Miniimagenet} &PN &Cosine&$43.23\pm 0.24\%$&$51.29\pm 0.22\%$&$26.99\pm 0.13\%$&$32.57\pm 0.12\%$  \\
&L-PN &Cosine&$\mathbf{50.10\pm 0.25\%}$&$\mathbf{66.94\pm 0.21\%}$&$\mathbf{33.51\pm 0.15\%}$&$\mathbf{50.86\pm 0.14\%}$\\
\cmidrule[0.01pt]{2-7}
&PN &Euclid.&$47.98\pm 0.25\%$&$66.72\pm 0.21\%$&$31.91\pm 0.15 \%$&$51.50\pm 0.14 \%$ \\
&L-PN &Euclid.&$\mathbf{49.47\pm 0.25\%}$&$\mathbf{66.83\pm 0.21\%}$&$\mathbf{32.60\pm 0.15\%}$&$\mathbf{51.72\pm 0.14\%}$\\
\bottomrule
\end{tabular}
}
\caption{Few-shot learning on Omniglot and Mini-Imagenet with PN and L-PN. Results are averaged with $95\%$ confidence intervals. '$-$': not reported.}
\label{table: accuracy on PN}
\end{table*}

\textbf{Results on Few-Shot Learning.} The experiments are also conducted on Omniglot and Mini-Imagenet, and the parameter setup is the same as in Section \ref{sec:GNN-experiment}. The results are shown in Table \ref{table: accuracy on PN}. On Omniglot, L-PN performs comparably with PN. On Mini-Imagenet, L-PN consistently improves PN on \emph{every} learning task. For PN with Euclidean distance, the improvement on 1-shot learning is more significant than on 5-shot learning, which is because taking the average of multiple support samples in the same class helps PN learn a more discriminative embedding. 

For PN with cosine distance, L-PN improves PN by a huge margin. Originally, cosine PN performs much worse than Euclidean PN. Incorporating the large margin loss greatly boosts its performance and makes it comparable with or even better than Euclidean PN. This shows that the large margin distance loss function helps cosine PN learn a much better embedding space and presumably alleviate the gradient vanishing problem in training. 

\begin{figure}[!t]
	\centering
	\subfigure[Test accuracy as $m$ changes.]{
		\includegraphics[width=0.8\textwidth]{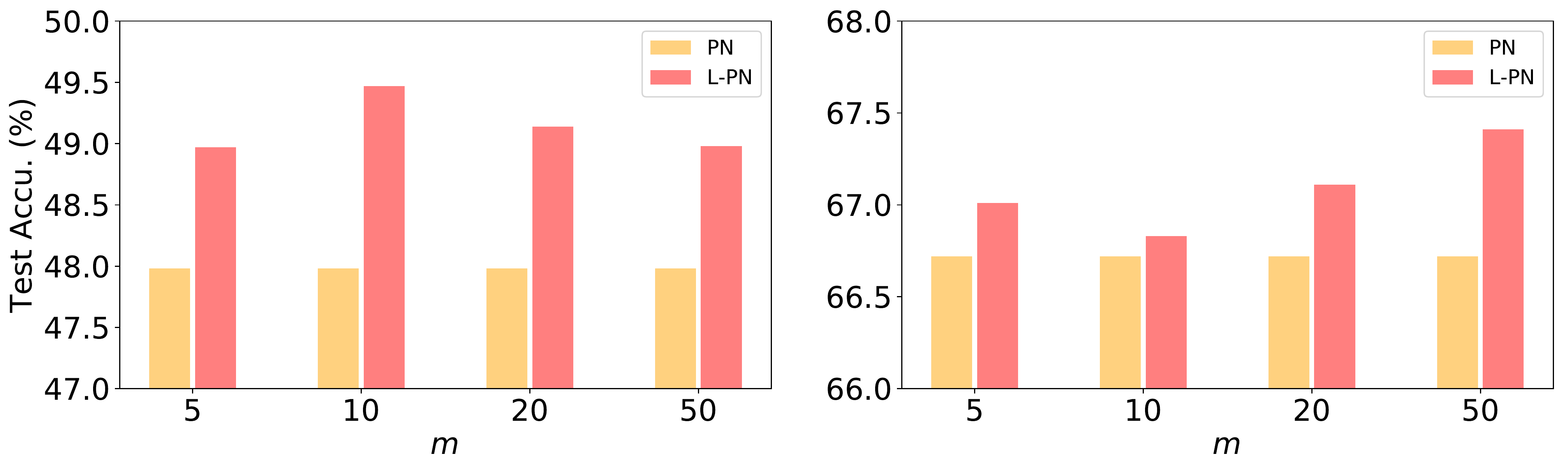}
		\label{fig: pn-margin}}
	\subfigure[Test accuracy as $\lambda$ changes.]{
		\includegraphics[width=0.8\textwidth]{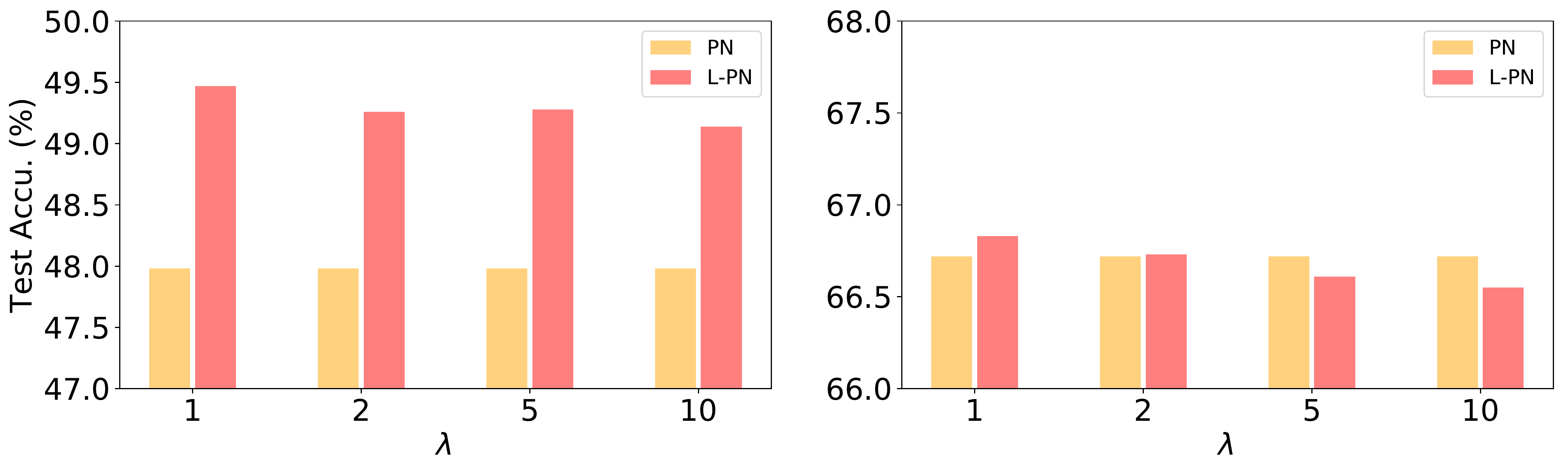}
		\label{fig: pn-lambda}}
    \captionsetup{justification=centering}

	\caption{$5$-way learning using PN on Mini-Imagenet. (Left: $1$-shot, right: $5$-shot)}
	\label{fig: robustness pn}
\end{figure}

\textbf{Parameter Sensitivity.} We also study the sensitivity of the balancing parameter $\lambda$ and the margin $m$ in L-PN. The results are shown in Fig. \ref{fig: robustness pn}. We can see that for $1$-shot learning, L-PN consistently outperforms PN as $\lambda$ and $m$ changes. For $5$-shot learning, L-PN outperforms PN as the margin $m$ changes from 5 to 50. It is only slightly worse than PN when the balancing parameter $\lambda$ is larger than 5. Overall, the results show that L-PN is robust. 

\textbf{Running Time.} The computational overhead of L-PN is very small. For $5$-way $5$-shot learning on Mini-Imagenet, one update of L-PN takes $0.109$s versus $0.105$s of PN, which only incurs $3.8\%$ computational overhead.

\section{Related Works} \label{sec:related work}
\subsection{Few-Shot Learning} 

Early work on few-shot learning focused on using generative models \cite{Feifei06} and inference methods \cite{Lake11}. With the recent success of deep learning, few-shot learning has been studied heavily with deep models and has made encouraging progress. Methods for few-shot learning can be roughly categorized as metric based methods, learning to fine-tune methods, and sequence based methods. 

\textbf{Metric Based Methods.} The basic idea of metric based methods is to learn a metric to measure the similarity between samples \cite{Vinyals16,Sung18}. \citet{Koch15} proposed siamese neural networks for one-shot learning. It learns a network which employs a unique structure to naturally rank similarity between inputs. \citet{Mehrotra17} proposed to use residual blocks to improve the expressive ability of siamese networks. It argues that having a learnable and more expressive similarity objective is an essential part for few-shot learning. \citet{Bertinetto16} proposed to learn the parameters of a deep model (pupil network) in one shot by constructing a second neural network which predicts the parameters of the pupil network from a single sample. To make this approach feasible, it proposed a number of factorizations of the parameters of the pupil network.
\citet{Vinyals16} proposed to learn a matching network, which maps a small labeled support set and an unlabeled example to its label by using a recurrent neural network. This network employs attention and memory mechanisms to enable rapid learning and it is based on a principle: test and train conditions must match. It proposed an episode-based training procedure for few-shot learning which has been followed by many papers. \citet{Snell17} proposed prototypical networks to do few-shot classification by computing distances to prototype representations of each class. The key idea is that the prototype is computed by taking the average of embedding vectors of samples from the same class. It was extended to Gaussian prototypical networks by \citet{Fort17}. \citet{Ren18} used prototypical networks to do semi-supervised few-shot learning. \citet{Kaiser17} proposed to achieve few-shot learning and life-long learning through continuous updating of memory in the learning process. It employs a large-scale memory module which uses fast nearest-neighbor algorithms. It can achieve lifelong learning without resetting the memory module during training. \citet{Garcia18} proposed graph neural networks for few-shot classification. It uses graph structure to model the relation between samples and can be extended to semi-supervised few-shot learning and active learning. \citet{Sung18} argued that the embedding space should be classified by a non-linear classifier and proposed to compare a support set and a query using a relation network where a distance criterion is learned via a trainable neural network to measure the similarity between two samples. Our large margin method can be potentially applied to almost all these models.

\textbf{Learning to Fine-Tune Methods.} \citet{Munkhdalai17} proposed meta networks which learn meta-level knowledge across tasks and can produce a new model through fast parameterization. 
\citet{Ravi17} proposed an LSTM-based meta-learner model by learning both an initial condition and a general optimization strategy for few-shot learning, which can be used to  update the learner network (classifier) in testing.
\citet{Finn17} proposed a model-agnostic meta-learning (MAML) approach which learns the initialization of a model, and based on this initialization, the model can quickly adapt to new tasks with a small number of gradient steps. It can be incorporated into many learning problems, such as classification, regression, and reinforcement learning. \citet{Li17} proposed Meta-SGD which learns not only the initialization, but also the update direction and the learning rate of stochastic gradient descent algorithms. Experiments show that it can learn faster and more accurately than MAML.
   
\textbf{Sequence Based Methods.} Sequence based methods for few-shot learning accumulate knowledge learned in the past and enable generalization on new samples with the learned knowledge. \citet{Santoro16} introduced an external memory on recurrent neural networks to make predictions with only a few samples. With the external memory, it offers the ability to quickly encode and retrieve new information. \citet{Mishra18} proposed a meta-learner architecture which uses temporal convolution and attention mechanism to accumulate past information. It can quickly incorporate past experience and can be applied to both few-shot learning and reinforcement learning.

\subsection{Large Margin Learning} 

Large margin methods \cite{Platt00,Tsochantaridis05,Weinberger05,Zien05,Parameswaran10,Zhang11,Schroff15} have been widely used in machine learning, including multiclass classification \cite{Platt00}, multi-task learning \cite{Parameswaran10}, transfer learning \cite{Zhang11}, etc. Due to the vast literature on large margin methods, we only review the most relevant works.
\citet{Weinberger05} proposed to learn a Mahalanobis distance metric for $k$-nearest neighbor classification by using semidefinite programming. The main idea is that $k$-nearest neighbors always belong to the same class and samples from different classes should be separated by a large margin. 
\citet{Parameswaran10} extended the large margin nearest neighbor algorithm to the multi-task paradigm. 
\citet{Schroff15} proposed a large margin method for face recognition by using the triplet loss to learn a mapping from face images to a compact Euclidean space. 
\citet{Zien05} proposed to maximize Jason-Shannon divergence for large margin nonlinear embeddings. The idea is to learn the embeddings of data with fixed decision boundaries, which is the opposite process of common classification methods. Our method is in spirit similar to this method in adding a large margin prior to learn the embeddings. 
There are also some works applying large margin methods for attribute-based zero-shot \cite{Akata13} and few-shot learning \cite{LiXin15}, but their problem setups are very different with the few-shot (meta) learning considered in this paper.

A number of recent works \cite{Ranjan17,Hadsell06,Sun14,Liu16,Liu17,Wang18,WangHao18,Deng18,LiuYu17} realize large margin embedding by defining various loss functions. \citet{Hadsell06} first proposed the contrastive loss and applied it to dimensionality reduction.
\citet{Sun14} combined the cross entropy loss and the contrastive loss to learn deep face representation. It reduces intra-personal variations and enlarges inter-personal differences by combining the identification task and the verification task. \citet{Liu16} proposed the large margin softmax loss for training convolutional neural networks. It explicitly encourages inter-class separability and intra-class compactness between embeddings by defining an adjustable and multiplicative margin. Motivated by that the learned features should be both discriminative and polymerized, \citet{LiuYu17} introduced the congenerous cosine algorithm to optimize the cosine similarity among data. \citet{Wang17} proposed the normface loss by introducing a vector for each class and optimizing the cosine similarity. \citet{WangHao18} proposed the cosface loss by defining an additive margin on the cosine space of $L_2$ normalized embeddings and weight vectors. \citet{Deng18} extended the cosface loss to the arcface loss by setting an additive margin on the angular space instead of the cosine space. All these loss functions can be applied to implement the large margin prior for few-shot learning. We discuss and compare these methods in the next section.

\section{Discussion}
\label{sec:discussion}

\begin{table*}[t]
	\centering
	\scalebox{1.0}{
		\begin{tabular}{cccc}
			\toprule
			& &\multicolumn{2}{c}{5-Way}\\
			Model&Dist&1-shot&5-shot\\
			\midrule
			PN &Euclid.&$47.98\pm 0.25\%$&$66.72\pm 0.21\%$ \\
			L-PN ($m=10$) &Euclid.&$49.47\pm 0.25\%$&$66.83\pm 0.21\%$\\
			PN+normalized triplet &Euclid.&$50.30\pm 0.25\%$&$67.13\pm 0.21\%$ \\
			\midrule
			GNN & $-$ &$50.33\pm 0.36 \%$&$66.41\pm 0.63\%$\\
			L-GNN ($m=10$) &$-$&$51.08\pm 0.69\%$&$67.57\pm 0.66\%$\\
			GNN+normalized triplet &$-$&$50.73\pm 0.70\%$&$67.28\pm 0.64\%$ \\
			GNN+normalized contrastive &$-$&$50.81\pm 0.69\%$&$67.49\pm 0.64\%$ \\
			GNN+normface &$-$&$51.41\pm 0.68\%$&$67.81\pm 0.64\%$\\
			GNN+cosface ($m=0.2$) &$-$&$51.49\pm 0.69\%$&$66.72\pm 0.65\%$\\
			GNN+arcface ($m=0.1$) &$-$&$\times$&$67.21\pm 0.64\%$\\
			\bottomrule
		\end{tabular}
	}
	\caption{$5$-way few-shot learning on Mini-Imagenet. Results are averaged with $95\%$ confidence intervals. '$-$': not reported. '$\times$': fail to converge.}
	\label{table: accuracy on $5$-way Mini-Imagenet different models}
\end{table*}

In this section, we implement and compare several of the aforementioned loss functions for large margin few-shot learning, including the normalized triplet loss, the normalized contrastive loss \cite{Hadsell06,Sun14}, the normface loss \cite{Wang17}, the cosface loss \cite{WangHao18}, and the arcface loss \cite{Deng18}. We test these models for $1$-shot and $5$-shot learning on Mini-Imagenet, and the results are summarized in Tables \ref{table: accuracy on $5$-way Mini-Imagenet different models}, \ref{table: accuracy on $5$-way Mini-Imagenet hyperparameter3}, \ref{table: accuracy on $5$-way Mini-Imagenet hyperparameter}, \ref{table: accuracy on $5$-way Mini-Imagenet hyperparameter2}.

All these models consider $L_2$ normalized weight vectors of the softmax classifier and the embeddings:
$$ \tilde{\mathbf{w}}_j = \frac{\mathbf{w}_j}{\parallel \mathbf{w}_j \parallel},~~~~~~~~ \tilde{f}_\phi(\mathbf{x}_i) = \frac{f_\phi(\mathbf{x}_i)}{\parallel f_\phi(\mathbf{x}_i) \parallel}.$$
After normalization, the cosine value of the angle between the $j$-th weight vector $\tilde{\mathbf{w}}_j$ and the embedding vector $\tilde{f}_\phi(\mathbf{x}_i)$ is $\cos(\theta_{j,i})=\tilde{\mathbf{w}}_j^T\tilde{f}_\phi(\mathbf{x}_i).$
By introducing a scale factor $s$, the softmax loss can be rewritten as:
\begin{equation}
\begin{split}
\label{eq. softmax loss}
\mathcal{L}_{\text{softmax}} = \frac{1}{N}\sum_{i}-\log \frac{e^{s\cos(\theta_{y_i,i})}}{\sum_{j}e^{s\cos(\theta_{j,i})}}.
\end{split}
\end{equation}

\textbf{Normalized triplet loss.}
Similarly, the normalized triplet loss can be defined as:
\begin{equation}
\begin{split}
\label{eq. normalized triplet loss}
\mathcal{L}_{\text{large-margin}}=\frac{1}{N_t}\sum_{i=1}^{N_t}\left[\parallel \tilde{f}_\phi(\mathbf{x}_i^a)-\tilde{f}_\phi(\mathbf{x}_i^p)\parallel_2^2- \parallel \tilde{f}_\phi(\mathbf{x}_i^a)-\tilde{f}_\phi(\mathbf{x}_i^n)\parallel_2^2+m\right]_+,
\end{split}
\end{equation}
where $m$ is the margin and $N_t$ is the number of triplets. The normalized triplet loss \eqref{eq. normalized triplet loss} can be combined with \eqref{eq. softmax loss} to train large margin graph neural networks (GNN) and the total training loss is:
\begin{equation}
\begin{split}
\label{eq. total training loss}
\mathcal{L} = \mathcal{L}_{\text{softmax}}+\lambda*\mathcal{L}_{\text{large-margin}}.
\end{split}
\end{equation}
Similarly, it can also be incorporated into the softmax loss to train large margin prototypical networks (PN), with the normalized embeddings scaled by a factor $s$.

We test GNN and PN with the augmented normalized triplet loss on Mini-Imagenet. For experiments in Table \ref{table: accuracy on $5$-way Mini-Imagenet different models}, we set the margin $m=0.5$, which is computed by $\frac{1}{2N_b}\sum_{i=1}^{N_b} \parallel f_\phi(\mathbf{x}_i)\parallel _2$, the average of the $L_2$-norm of all embeddings in one mini-batch with randomly initialized  parameters at the start of training, where $N_b$ is the number of all samples in one mini-batch. We set $\lambda=1$ for all experiments, and $s=10$ following \citet{Wang17}. We also test the normalized triplet loss with different $m$, $\lambda$, and $s$, as summarized in Table \ref{table: accuracy on $5$-way Mini-Imagenet hyperparameter}.
The results show that the normalized triplet loss consistently improves PN and GNN for all learning tasks, demonstrating the benefits of the large margin principle for few-shot learning. 
For PN, the normalized triplet loss performs even better than the unnormalized triplet loss. This shows that under some circumstances, normalizing the embedding space helps train a better classifier. However, for GNN, the normalized triplet loss is not as robust as the unnormalized triplet loss. As shown in Tables \ref{table: accuracy on $5$-way Mini-Imagenet hyperparameter3} and \ref{table: accuracy on $5$-way Mini-Imagenet hyperparameter}, the latter consistently outperforms the former.

\begin{table*}[t]
	\centering
	\resizebox{0.45\textwidth}{!}{
		\begin{tabular}{|c|c|c|c|}  
			\toprule
			&&&L-GNN \\
			\midrule 
			\multirow{4}{*}{$1$-shot}&\multirow{2}{*}{$\lambda=1$}&\multirow{1}{*}{$m=10$}&$51.08\pm 0.69\%$ \\
			&&\multirow{1}{*}{$m=20$}&$51.42\pm 0.69\%$ \\
			
			&\multirow{2}{*}{$\lambda=2$}&\multirow{1}{*}{$m=10$}&$51.04\pm 0.71\%$ \\
			&&\multirow{1}{*}{$m=20$}&$51.12\pm 0.69\%$ \\
			\cmidrule{2-4} 
			
			&\multicolumn{2}{c|}{GNN}&$50.33\pm 0.36\%$\\
			
			\midrule
			
			\multirow{4}{*}{$5$-shot}&\multirow{2}{*}{$\lambda=1$}&\multirow{1}{*}{$m=10$}&$67.57\pm 0.66\%$ \\
			&&\multirow{1}{*}{$m=20$}&$67.86\pm 0.65\%$ \\
			
			&\multirow{2}{*}{$\lambda=2$}&\multirow{1}{*}{$m=10$}&$67.59\pm 0.65\%$ \\
			&&\multirow{1}{*}{$m=20$}&$67.61\pm 0.64\%$ \\
			\cmidrule{2-4} 
			
			&\multicolumn{2}{c|}{GNN}&$66.41\pm 0.63\%$\\
			\bottomrule
		\end{tabular}
	}
	\caption{$5$-way few-shot learning using L-GNN (with the unnormalized triplet loss) on Mini-Imagenet. Results are averaged with $95\%$ confidence intervals.}
	\label{table: accuracy on $5$-way Mini-Imagenet hyperparameter3}
\end{table*}

\textbf{Normalized contrastive loss.}
The contrastive loss was first introduced by \citet{Hadsell06} and is defined as:
\begin{equation}
\begin{split}
\label{eq. unnormalized contrastive loss}
\mathcal{L}_{\text{largin-margin}}= \begin{cases}
 \parallel f_\phi(\mathbf{x}_i)-f_\phi(\mathbf{x}_j)\parallel_2^2& y_i=y_j, \\
 \max(0, m-\parallel f_\phi(\mathbf{x}_i)-f_\phi(\mathbf{x}_j)\parallel_2^2)& y_i\neq y_j, \\
 \end{cases} 
\end{split}
\end{equation}
where $m$ is the margin. The idea is to pull neighbors together and push non-neighbors apart.
The normalized contrastive loss can be defined similarly by replacing the embeddings ($f_\phi(\mathbf{x}_i)$, $f_\phi(\mathbf{x}_j)$) with the normalized ones ($\tilde{f}_\phi(\mathbf{x}_i)$, $\tilde{f}_\phi(\mathbf{x}_j)$).
Similar to the normalized triplet loss, the normalized contrastive loss can also be combined with the softmax loss to train large margin PN and large margin GNN.

We test GNN with the augmented normalized contrastive loss on Mini-Imagenet.  We can see that the normalized contrastive loss consistently outperforms GNN for all learning tasks, which again confirms that the large margin principle is useful for few-shot learning. The results are summarized in Tables \ref{table: accuracy on $5$-way Mini-Imagenet different models} and \ref{table: accuracy on $5$-way Mini-Imagenet hyperparameter}, where the parameter setup is the same as the normalized triplet loss. We can see from Table \ref{table: accuracy on $5$-way Mini-Imagenet hyperparameter3} and \ref{table: accuracy on $5$-way Mini-Imagenet hyperparameter} that the normalized contrastive loss is comparable to the unnormalized triplet loss for $5$-shot learning, but not as robust for the more challenging $1$-shot learning. We have also tested the unnormalized contrastive loss in our experiments, but find that it is very unstable and easy to diverge in training.

\textbf{Normface loss.}
The normface loss \cite{Wang17} was proposed to improve the performance of face verification. It identifies and studies the issues related to applying $L_2$ normalization on the embeddings and the weight vectors of the softmax classifier. Four different kinds of loss functions were proposed by \citet{Wang17}, and here we use the best model reported in \cite{Wang17}. The normface loss consists of two parts. One is the softmax loss \eqref{eq. softmax loss}, and the other part is another form of the contrastive loss:
\begin{equation}
\begin{split}
\label{eq. normface loss}
\mathcal{L}_{\text{largin-margin}}= \begin{cases}
 \parallel \tilde{f}_\phi(\mathbf{x}_i)-\tilde{\mathbf{w}}_j\parallel_2^2& y_i=j, \\
 \max(0, m-\parallel \tilde{f}_\phi(\mathbf{x}_i)-\tilde{\mathbf{w}}_j\parallel_2^2)& y_i\neq j, \\
 \end{cases}
\end{split}
\end{equation}
which is obtained by replacing $\tilde{f}_\phi(\mathbf{x}_j)$ in the normalized contrastive loss with the normalized weight vector $\tilde{\mathbf{w}}_j$ of the softmax classifier. 
The normface loss can also be combined with the softmax loss \eqref{eq. softmax loss} to train large margin GNN. However, it can not be directly applied on PN because PN uses a non-parametric classifier. 

We test GNN with the augmented normface loss on Mini-Imagenet. The experimental setup is the same as the normalized triplet loss. Results in Table \ref{table: accuracy on $5$-way Mini-Imagenet different models}, \ref{table: accuracy on $5$-way Mini-Imagenet hyperparameter3} and \ref{table: accuracy on $5$-way Mini-Imagenet hyperparameter} show that the normface loss significantly outperforms GNN for all learning tasks, and is robust and comparable to the unnormalized triplet loss. This suggests that the normface loss may also be a good alternative to implement the large margin principle for metric based few-shot learning methods which use a parametric classifier.

\begin{table*}[t]
	\centering
	\resizebox{\textwidth}{!}{
		\begin{tabular}{|c|c|c|c|c|c|c|}  
			\toprule
			&&&&Normalized triplet loss& Normalized contrastive loss&Normface loss \\
			\midrule 
			\multirow{8}{*}{$1$-shot}&\multirow{4}{*}{$s=10$}&\multirow{2}{*}{$\lambda=1$}&\multirow{1}{*}{$m=0.5$}&$50.73\pm 0.70\%$&$50.81\pm 0.69\%$&$51.41\pm 0.68\%$ \\
			&&&\multirow{1}{*}{$m=1.0$}&$50.88\pm 0.71\%$&$50.67\pm 0.71\%$&$51.60\pm 0.69\%$ \\
			
			&&\multirow{2}{*}{$\lambda=2$}&\multirow{1}{*}{$m=0.5$}&$50.98\pm 0.69\%$&$50.53\pm 0.70\%$&$51.35\pm 0.71\%$ \\
			&&&\multirow{1}{*}{$m=1.0$}&$51.06\pm 0.70\%$&$50.68\pm 0.70\%$&$51.15\pm 0.71\%$\\
			
			&\multirow{4}{*}{$s=20$}&\multirow{2}{*}{$\lambda=1$}&\multirow{1}{*}{$m=0.5$}&$50.59\pm 0.70\%$&$51.13\pm 0.71\%$&$51.37\pm 0.73\%$ \\
			&&&\multirow{1}{*}{$m=1.0$}&$50.48\pm 0.70\%$&$50.67\pm 0.68\%$&$51.32\pm 0.70\%$\\
			&&\multirow{2}{*}{$\lambda=2$}&\multirow{1}{*}{$m=0.5$}&$51.12\pm 0.70\%$&$51.40\pm 0.69\%$&$51.57\pm 0.69\%$ \\
			&&&\multirow{1}{*}{$m=1.0$}&$50.61\pm 0.70\%$&$51.38\pm 0.71\%$&$51.12\pm 0.68\%$\\
			\cmidrule{2-7} 
			
			&\multicolumn{3}{c|}{GNN}&$50.33\pm 0.36\%$&$50.33\pm 0.36\%$&$50.33\pm 0.36\%$\\
			\midrule
			
			\multirow{8}{*}{$5$-shot}&\multirow{4}{*}{$s=10$}&\multirow{2}{*}{$\lambda=1$}&\multirow{1}{*}{$m=0.5$}&$67.28\pm 0.64\%$&$67.49\pm 0.64\%$&$67.81\pm 0.64\%$ \\
			&&&\multirow{1}{*}{$m=1.0$}&$67.31\pm 0.64\%$&$67.10\pm 0.67\%$&$67.25\pm 0.64\%$\\
			
			&&\multirow{2}{*}{$\lambda=2$}&\multirow{1}{*}{$m=0.5$}&$67.47\pm 0.66\%$&$67.55\pm 0.66\%$&$67.28\pm 0.65\%$ \\
			&&&\multirow{1}{*}{$m=1.0$}&$66.91\pm 0.64\%$&$67.74\pm 0.65\%$&$67.44\pm 0.64\%$\\
			&\multirow{4}{*}{$s=20$}&\multirow{2}{*}{$\lambda=1$}&\multirow{1}{*}{$m=0.5$}&$67.09\pm 0.66\%$&$67.05\pm 0.64\%$&$67.51\pm 0.63\%$ \\
			&&&\multirow{1}{*}{$m=1.0$}&$66.75\pm 0.64\%$&$67.18\pm 0.65\%$&$67.58\pm 0.64\%$\\
			&&\multirow{2}{*}{$\lambda=2$}&\multirow{1}{*}{$m=0.5$}&$67.20\pm 0.65\%$&$67.35\pm 0.66\%$&$67.76\pm 0.64\%$ \\
			&&&\multirow{1}{*}{$m=1.0$}&$67.09\pm 0.67\%$&$67.68\pm 0.64\%$&$67.87\pm 0.65\%$\\
			\cmidrule{2-7} 
			
			&\multicolumn{3}{c|}{GNN}&$66.41\pm 0.63\%$&$66.41\pm 0.63\%$&$66.41\pm 0.63\%$\\
			\bottomrule
		\end{tabular}
	}
	\caption{$5$-way few-shot learning using GNN with various loss functions on Mini-Imagenet. Results are averaged with $95\%$ confidence intervals.}
	\label{table: accuracy on $5$-way Mini-Imagenet hyperparameter}
\end{table*}

\textbf{Cosface loss.}
\citet{WangHao18} proposed the cosface loss, which is defined as:
\begin{equation}
\begin{split}
\label{eq. cosface loss}
\mathcal{L} =\mathcal{L}_{\text{largin-margin}}
 = \frac{1}{N}\sum_{i}-\log \frac{e^{s(\cos(\theta_{y_i,i})-m)}}{e^{s(\cos(\theta_{y_i,i})-m)}+\sum_{j\neq y_i}e^{s\cos(\theta_{j,i})}}.
\end{split}
\end{equation}
It introduces a margin in the cosine space of  $L_2$ normalized embeddings and weight vectors of the softmax classifier, which can render the learned features more discriminative. 
The cosface loss can be applied to train large margin GNN. However, it can not be directly applied on PN as PN uses a non-parametric classifier. 

We test GNN with the cosface loss function on Mini-Imagenet. For the experiments in Table \ref{table: accuracy on $5$-way Mini-Imagenet different models}, we set the margin $m=0.2$. We also test with different $m$ in Table \ref{table: accuracy on $5$-way Mini-Imagenet hyperparameter2}. For all the experiments, we set $s=10$ as in the normalized triplet loss.
Results in Table \ref{table: accuracy on $5$-way Mini-Imagenet different models} show that the cosface loss can perform better than GNN when $m$ is chosen properly, which again shows the usefulness of the large margin principle. However, the selection of $m$ is non-trivial. It was suggested in \cite{WangHao18} that for face recognition, the proper choice of $m$ is $0\sim0.45$. However, for few-shot learning, as shown in Table \ref{table: accuracy on $5$-way Mini-Imagenet hyperparameter2}, when $m$ increases, the performance of the cosface loss decreases significantly. This shows that the cosface loss is sensitive to the margin, and overall is not comparable to the unnormalized triplet loss.

\textbf{Arcface loss.}
\citet{Deng18} proposed the arcface loss, which is defined as:
\begin{equation}
\begin{split}
\label{eq. arcface loss}
\mathcal{L} =\mathcal{L}_{\text{largin-margin}} = \frac{1}{N}\sum_{i}-\log \frac{e^{s(\cos(\theta_{y_i,i}+m))}}{e^{s(\cos(\theta_{y_i,i}+m))}+\sum_{j\neq y_i}e^{s\cos(\theta_{j,i})}}.
\end{split}
\end{equation}
The arcface loss extends the cosface loss by defining the margin $m$ in the angular space instead of the cosine space. The angular margin has a clearer geometric interpretation than the cosine margin. It was reported in \cite {Deng18} that the arcface loss can obtain more discriminative deep features compared to other multiplicative angular margin and additive cosine margin methods. Similar to the cosface loss, it can be applied to train large margin GNN, but can not be directly applied on PN. 

We test GNN with the arcface loss function on Mini-Imagenet. For the experiments in Table \ref{table: accuracy on $5$-way Mini-Imagenet different models}, we set the margin $m=0.1$. We also test with different $m\in [0, 0.5]$ as suggested by \cite{Deng18} in Table \ref{table: accuracy on $5$-way Mini-Imagenet hyperparameter2}. For all the experiments, we set $s=10$ as in the normalized triplet loss.
The results show that the arcface loss can perform better than GNN with $m=0.1$ for $5$-shot learning, which again confirms the effectiveness of the large margin principle. However, as shown in Table \ref{table: accuracy on $5$-way Mini-Imagenet hyperparameter2}, the arcface loss diverges in training for most cases tested, and only converges in one case. This shows that the arcface loss is very sensitive to $m$, and is not comparable to the unnormalized triplet loss for few-shot learning.

To summarize, the experiments show that all large margin losses can substantially improve the original few-shot learning model, which demonstrates the benefits of the large margin principle. Compared with other loss functions, the unnormalized triplet loss has two clear advantages. First, it is more general and can be easily incorporated into metric based few-shot learning methods. As mentioned above, the normface loss, the cosface loss, and the arcface loss can not be directly applied to few-shot learning methods using non-parametric classifiers. Second, it is more robust than other loss functions such as the normalized triplet loss, the normalized contrastive loss, the cosface loss, and the arcface loss. 
On the running time, the computational overheads of these loss functions are all small, similar to that of the unnormalized triplet loss.

\begin{table*}[t]
	\centering
	\resizebox{0.8\textwidth}{!}{
		\begin{tabular}{|c|c|c|c|c|}  
			\toprule
			\multirow{2}{*}{$m$}&\multicolumn{2}{c}{Cosface loss}& \multicolumn{2}{|c|}{Arcface loss} 
			\\
			\cmidrule{2-5}
			&$1$-shot& $5$-shot&$1$-shot&$5$-shot \\
			\midrule
			$0.1$&$51.04\pm 0.69\%$&$67.04\pm 0.64\%$&$\times$&$67.21\pm 0.64\%$ \\
			$0.2$&$51.49\pm 0.69\%$&$66.72\pm 0.65\%$&$\times$&$\times$ \\
			$0.3$&$50.91\pm 0.69\%$&$66.30\pm 0.65\%$&$\times$&$\times$ \\
			$0.4$&$50.55\pm 0.68\%$&$65.48\pm 0.67\%$&$\times$&$\times$ \\
			$0.5$&$50.46\pm 0.69\%$&$64.93\pm 0.65\%$&$\times$&$\times$ \\
			\cmidrule{1-5} 
			
			GNN&$50.33\pm 0.36\%$&$66.41\pm 0.63\%$&$50.33\pm 0.36\%$&$66.41\pm 0.63\%$\\
			\bottomrule
		\end{tabular}
	}
	\caption{$5$-way few-shot learning using GNN with various loss functions on Mini-Imagenet. Results are averaged with $95\%$ confidence intervals. '$\times$': fail to converge.}
	\label{table: accuracy on $5$-way Mini-Imagenet hyperparameter2}
\end{table*}

\section{Conclusions} \label{sec:conclusion}

This paper proposes a large margin principle for metric based few-shot learning and demonstrates its usefulness in improving the generalization capacity of two state-of-the-art methods. Our framework is simple, efficient, robust, and can be applied to benefit many existing and future few-shot learning methods. Future work includes developing theoretical guarantees for large margin few-shot learning and applying our method to solve real-word problems.

\bibliographystyle{iclr2018_conference}
\bibliography{named}

\begin{thebibliography}{43}
\providecommand{\natexlab}[1]{#1}
\providecommand{\url}[1]{\texttt{#1}}
\expandafter\ifx\csname urlstyle\endcsname\relax
  \providecommand{\doi}[1]{doi: #1}\else
  \providecommand{\doi}{doi: \begingroup \urlstyle{rm}\Url}\fi

\bibitem[Akata et~al.(2013)Akata, Perronnin, Harchaoui, and Schmid]{Akata13}
Zeynep Akata, Florent Perronnin, Zaid Harchaoui, and Cordelia Schmid.
\newblock Label-embedding for attribute-based classification.
\newblock In CVPR, 2013.

\bibitem[Bertinetto et~al.(2016)Bertinetto, Henriques, Valmadre, and
  Vedaldi]{Bertinetto16}
Luca Bertinetto, João~F. Henriques, Jack Valmadre, and Philip H. S.
  Torrand~Andrea Vedaldi.
\newblock Learning feed-forward one-shot learners.
\newblock In NIPS, 2016.

\bibitem[Deng et~al.(2018)Deng, Guo, and Zafeiriou]{Deng18}
Jiankang Deng, Jia Guo, and Stefanos Zafeiriou.
\newblock Arcface: Additive angular margin loss for deep face recognition.
\newblock In arXiv:1801.07698, 2018.

\bibitem[Fei-Fei et~al.(2006)Fei-Fei, Fergus, and Perona]{Feifei06}
Li~Fei-Fei, Rob Fergus, and Pietro Perona.
\newblock One-shot learning of object categories.
\newblock In TPAMI, 2006.

\bibitem[Finn et~al.(2017)Finn, Abbeel, and Levine]{Finn17}
Chelsea Finn, Pieter Abbeel, and Sergey Levine.
\newblock Model-agnostic meta-learning for fast adaptation of deep networks.
\newblock In ICML, 2017.

\bibitem[Fort(2017)]{Fort17}
Stanislav Fort.
\newblock Gaussian prototypical networks for few-shot learning on omniglot.
\newblock In NIPS workshop, 2017.

\bibitem[Fu et~al.(2015)Fu, Hospedales, Xiang, and Gong]{FuYanWei15}
Yanwei Fu, Timothy~M. Hospedales, Tao Xiang, and Shaogang Gong.
\newblock Transductive multi-view zero-shot learning.
\newblock In TPAMI, 2015.

\bibitem[Garcia \& Bruna(2018)Garcia and Bruna]{Garcia18}
Victor Garcia and Joan Bruna.
\newblock Few-shot learning with graph neural networks.
\newblock In ICLR, 2018.

\bibitem[Hadsell et~al.(2006)Hadsell, Chopra, and LeCun]{Hadsell06}
Raia Hadsell, Sumit Chopra, and Yann LeCun.
\newblock Dimensionality reduction by learning an invariant mapping.
\newblock In CVPR, 2006.

\bibitem[He et~al.(2016)He, Zhang, Ren, and Sun]{He16}
Kaiming He, Xiangyu Zhang, Shaoqing Ren, and Jian Sun.
\newblock Deep residual learning for image recognition.
\newblock In CVPR, 2016.

\bibitem[Kaiser et~al.(2017)Kaiser, Nachum, Roy, and Bengio]{Kaiser17}
{\L}ukasz Kaiser, Ofir Nachum, Aurko Roy, and Samy Bengio.
\newblock Learning to remember rare events.
\newblock In ICLR, 2017.

\bibitem[Koch et~al.(2015)Koch, Zemel, and Salakhutdinov]{Koch15}
Gregory Koch, Richard Zemel, and Ruslan Salakhutdinov.
\newblock Siamese neural networks for one-shot image recognition.
\newblock In ICML, 2015.

\bibitem[Krizhevsky et~al.(2012)Krizhevsky, Sutskever, and
  Hinton]{Krizhevsky12}
Alex Krizhevsky, Ilya Sutskever, and Geoffrey~E. Hinton.
\newblock Imagenet classification with deep convolutional neural networks.
\newblock In NIPS, 2012.

\bibitem[Lake et~al.(2011)Lake, Salakhutdinov, Gross, and Tenenbaum]{Lake11}
Brenden~M. Lake, Ruslan Salakhutdinov, Jason Gross, and Joshua~B. Tenenbaum.
\newblock One shot learning of simple visual concepts.
\newblock In CogSci, 2011.

\bibitem[Lampert et~al.(2009)Lampert, Nickisch, and Harmeling]{Lampert09}
Christoph~H. Lampert, Hannes Nickisch, and Stefan Harmeling.
\newblock Learning to detect unseen object classes by between-class attribute
  transfer.
\newblock In CVPR, 2009.

\bibitem[Li \& Guo(2015)Li and Guo]{LiXin15}
Xin Li and Yuhong Guo.
\newblock Max-margin zero-shot learning for multi-class classification.
\newblock In AISTATS, 2015.

\bibitem[Li et~al.(2017)Li, Zhou, Chen, and Li]{Li17}
Zhenguo Li, Fengwei Zhou, Fei Chen, and Hang Li.
\newblock Meta-sgd: Learning to learn quickly for few-shot learning.
\newblock In arXiv:1707.09835, 2017.

\bibitem[Liu et~al.(2016)Liu, Wen, Yu, and Yang]{Liu16}
Weiyang Liu, Yandong Wen, Zhiding Yu, and Meng Yang.
\newblock Large-margin softmax loss for convolutional neural networks.
\newblock In ICML, 2016.

\bibitem[Liu et~al.(2017{\natexlab{a}})Liu, Wen, Yu, Li, Raj, and Song]{Liu17}
Weiyang Liu, Yandong Wen, Zhiding Yu, Ming Li, Bhiksha Raj, and Le~Song.
\newblock Sphereface: Deep hypersphere embedding for face recognition.
\newblock In CVPR, 2017{\natexlab{a}}.

\bibitem[Liu et~al.(2017{\natexlab{b}})Liu, Li, and Wang]{LiuYu17}
Yu~Liu, Hongyang Li, and Xiaogang Wang.
\newblock Rethinking feature discrimination and polymerization for large-scale
  recognition.
\newblock In NIPS Workshop, 2017{\natexlab{b}}.

\bibitem[Mehrotra \& Dukkipati(2017)Mehrotra and Dukkipati]{Mehrotra17}
Akshay Mehrotra and Ambedkar Dukkipati.
\newblock Generative adversarial residual pairwise networks for one shot
  learning.
\newblock In arXiv:1703.08033, 2017.

\bibitem[Mishra et~al.(2018)Mishra, Rohaninejad, Chen, and Abbeel]{Mishra18}
Nikhil Mishra, Mostafa Rohaninejad, Xi~Chen, and Pieter Abbeel.
\newblock A simple neural attentive meta-learner.
\newblock In ICLR, 2018.

\bibitem[Mood \& Graybill(1974)Mood and Graybill]{Mood74}
Alexander~M. Mood and Franklin~A. Graybill.
\newblock Introduction to the theory of statistics.
\newblock In McGraw Hill: New York, 1974.

\bibitem[Munkhdalai \& Yu(2017)Munkhdalai and Yu]{Munkhdalai17}
Tsendsuren Munkhdalai and Hong Yu.
\newblock Meta networks.
\newblock In ICML, 2017.

\bibitem[Parameswaran \& Weinberger(2010)Parameswaran and
  Weinberger]{Parameswaran10}
Shibin Parameswaran and Kilian~Q. Weinberger.
\newblock Large margin multi-task metric learning.
\newblock In NIPS, 2010.

\bibitem[Platt et~al.(2000)Platt, Cristianini, and Shawe-Taylor]{Platt00}
John~C. Platt, Nello Cristianini, and John Shawe-Taylor.
\newblock Large margin dags for multiclass classification.
\newblock In NIPS, 2000.

\bibitem[Ranjan et~al.(2017)Ranjan, Castillo, and Chellappa]{Ranjan17}
Rajeev Ranjan, Carlos~D. Castillo, and Rama Chellappa.
\newblock L2-constrained softmax loss for discriminative face verification.
\newblock In arXiv:1703.09507, 2017.

\bibitem[Ravi \& Larochelle(2017)Ravi and Larochelle]{Ravi17}
Sachin Ravi and Hugo Larochelle.
\newblock Optimization as a model for few-shot learning.
\newblock In ICLR, 2017.

\bibitem[Ren et~al.(2018)Ren, Triantafillou, Ravia, Snell, Swersky, Tenenbaum,
  Larochelle, and Zemel]{Ren18}
Mengye Ren, Eleni Triantafillou, Sachin Ravia, Jake Snell, Kevin Swersky,
  Joshua~B. Tenenbaum, Hugo Larochelle, and Richard~S. Zemel.
\newblock Meta-learning for semi-supervised few-shot classification.
\newblock In ICLR, 2018.

\bibitem[Santoro et~al.(2016)Santoro, Bartunov, Botvinick, Wierstra, and
  Lillicrap]{Santoro16}
Adam Santoro, Sergey Bartunov, Matthew Botvinick, Daan Wierstra, and Timothy
  Lillicrap.
\newblock Meta-learning with memory-augmented neural networks.
\newblock In ICML, 2016.

\bibitem[Schroff et~al.(2015)Schroff, Kalenichenko, and Philbin]{Schroff15}
Florian Schroff, Dmitry Kalenichenko, and James Philbin.
\newblock Facenet: A unified embedding for face recognition and clustering.
\newblock In CVPR, 2015.

\bibitem[Snell et~al.(2017)Snell, Swersky, and Zemel]{Snell17}
J.~Snell, K.~Swersky, and R.~S. Zemel.
\newblock Prototypical networks for few-shot learning.
\newblock In NIPS, 2017.

\bibitem[Sun et~al.(2014)Sun, Wang, and Tang]{Sun14}
Yi~Sun, Xiaogang Wang, and Xiaoou Tang.
\newblock Deep learning face representation by joint
  identification-verification.
\newblock In NIPS, 2014.

\bibitem[Sung et~al.(2018)Sung, Yang, Zhang, Xiang, Torr, and
  Hospedales]{Sung18}
Flood Sung, Yongxin Yang, Li~Zhang, Tao Xiang, Philip H.~S. Torr, and
  Timothy~M. Hospedales.
\newblock Learning to compare: Relation network for few-shot learning.
\newblock In CVPR, 2018.

\bibitem[Tsochantaridis et~al.(2005)Tsochantaridis, Joachims, Hofmann, and
  Altun]{Tsochantaridis05}
Ioannis Tsochantaridis, Thorsten Joachims, Thomas Hofmann, and Yasemin Altun.
\newblock Large margin methods for structured and interdependent output
  variables.
\newblock In JMLR, 2005.

\bibitem[Vinyals et~al.(2016)Vinyals, Blundell, Lillicrap, and
  Wierstra]{Vinyals16}
O.~Vinyals, C.~Blundell, T.~Lillicrap, and D.~Wierstra.
\newblock Matching networks for one shot learning.
\newblock In NIPS, 2016.

\bibitem[Wang et~al.(2017)Wang, Xiang, Cheng, and Yuille]{Wang17}
Feng Wang, Xiang Xiang, Jian Cheng, and Alan~L. Yuille.
\newblock Normface: L2 hypersphere embedding for face verification.
\newblock In ACM MM, 2017.

\bibitem[Wang et~al.(2018{\natexlab{a}})Wang, Liu, Liu, and Cheng]{Wang18}
Feng Wang, Weiyang Liu, Haijun Liu, and Jian Cheng.
\newblock Additive margin softmax for face verification.
\newblock In arXiv:1801.05599, 2018{\natexlab{a}}.

\bibitem[Wang et~al.(2018{\natexlab{b}})Wang, Wang, Zhou, Ji, Gong, Zhou, Li,
  and Liu]{WangHao18}
Hao Wang, Yitong Wang, Zheng Zhou, Xing Ji, Dihong Gong, Jingchao Zhou, Zhifeng
  Li, and Wei Liu.
\newblock Cosface: Large margin cosine loss for deep face recognition.
\newblock In CVPR, 2018{\natexlab{b}}.

\bibitem[Weinberger et~al.(2005)Weinberger, Blitzer, and Saul]{Weinberger05}
Kilian~Q. Weinberger, John Blitzer, and Lawrence~K. Saul.
\newblock Distance metric learning for large margin nearest neighbor
  classification.
\newblock In NIPS, 2005.

\bibitem[Xian et~al.(2018)Xian, Lampert, Schiele, and Akata]{Xian18}
Yongqin Xian, Christoph~H. Lampert, Bernt Schiele, and Zeynep Akata.
\newblock Zero-shot learning - a comprehensive evaluation of the good, the bad
  and the ugly.
\newblock In TPAMI, 2018.

\bibitem[Zhang et~al.(2011)Zhang, He, Liu, Si, and Lawrence]{Zhang11}
Dan Zhang, Jingrui He, Yan Liu, Luo Si, and Richard~D. Lawrence.
\newblock Multi-view transfer learning with a large margin approach.
\newblock In KDD, 2011.

\bibitem[Zien \& Candela(2005)Zien and Candela]{Zien05}
Alexander Zien and Joaquin~Quinonero Candela.
\newblock Large margin non-linear embedding.
\newblock In ICML, 2005.

\end{thebibliography}

\end{document}